# Analysis of the first Genetic Engineering Attribution Challenge

Oliver M. Crook[1], Kelsey Lane Warmbrod[2,3], Greg Lipstein[4], Christine Chung[4], Christopher W. Bakerlee[5], T. Greg McKelvey Jr.[5], Shelly R. Holland[5], Jacob L. Swett[5], Kevin M. Esvelt[5,6], Ethan C. Alley[5,6,*], and William J. Bradshaw[5,6,†]

[1] Oxford Protein Informatics Group, Department of Statistics, University of Oxford, Oxford, United Kingdom
[2] Johns Hopkins Center for Health Security, Johns Hopkins Bloomberg School of Public Health, Baltimore, MD, USA
[3] Institute of Public Health Genetics, University of Washington, Seattle, WA, USA
[4] DrivenData Inc, Denver, CO, USA
[5] altLabs Inc, Berkeley, CA, USA
[6] Media Laboratory, Massachusetts Institute of Technology, Cambridge, MA, USA
[*] For correspondence: ethan@altlabs.tech
[†] For correspondence: wjbrad@mit.edu

## Abstract

The ability to identify the designer of engineered biological sequences – termed genetic engineering attribution (GEA) – would help ensure due credit for biotechnological innovation, while holding designers accountable to the communities they affect. Here, we present the results of the first Genetic Engineering Attribution Challenge, a public data-science competition to advance GEA. Top-scoring teams dramatically outperformed previous models at identifying the true lab-of-origin of engineered sequences, including an increase in top-1 and top-10 accuracy of 10 percentage points. A simple ensemble of prizewinning models further increased performance. New metrics, designed to assess a model's ability to confidently exclude candidate labs, also showed major improvements, especially for the ensemble. Most winning teams adopted CNN-based machine-learning approaches; however, one team achieved very high accuracy with an extremely fast neural-network-free approach. Future work, including future competitions, should further explore a wide diversity of approaches for bringing GEA technology into practical use.

## Introduction

Genetic engineering is becoming increasingly powerful, widespread, and accessible, enabling ever-more people to manipulate organisms in increasingly sophisticated ways. As biotechnology advances and spreads, the ability to attribute genetically engineered organisms to their designers becomes increasingly important – both as a means to ensure due recognition and prevent plagiarism, and as a means of holding these designers accountable to the communities their work affects[1–4]. While many academic researchers openly claim credit for their strains and sequences, the provenance of other products – including unpublished work, the products of industrial and government labs, and the work of amateur enthusiasts – is often more difficult to establish.

While tools for attributing these products of biotechnology – for *genetic engineering attribution* (GEA) – have historically lagged behind the pace of scientific development, recent years have seen rapid progress[1,2,5,6]. Genetic engineers face many design choices when creating an engineered nucleic-acid sequence, and the sum of these choices constitutes a design signature which, in at least some cases, is detectable by GEA algorithms[2,5] (Fig. 1a). The more reliably and precisely these algorithms can identify the true designer of a sequence, the greater the potential benefits for accountability and innovation.

Past work on GEA[2,5,6] has largely focused on predicting the origin lab of plasmid sequences from the Addgene data repository. Performance on this problem has improved rapidly (Fig. 1b). Most recently, Alley *et al.* used a Recurrent Neural Network (RNN) approach to achieve an accuracy of 70% and a top-10 accuracy (the frequency with which the true lab-of-origin is within the model's top-10 predictions) of 85%[2].

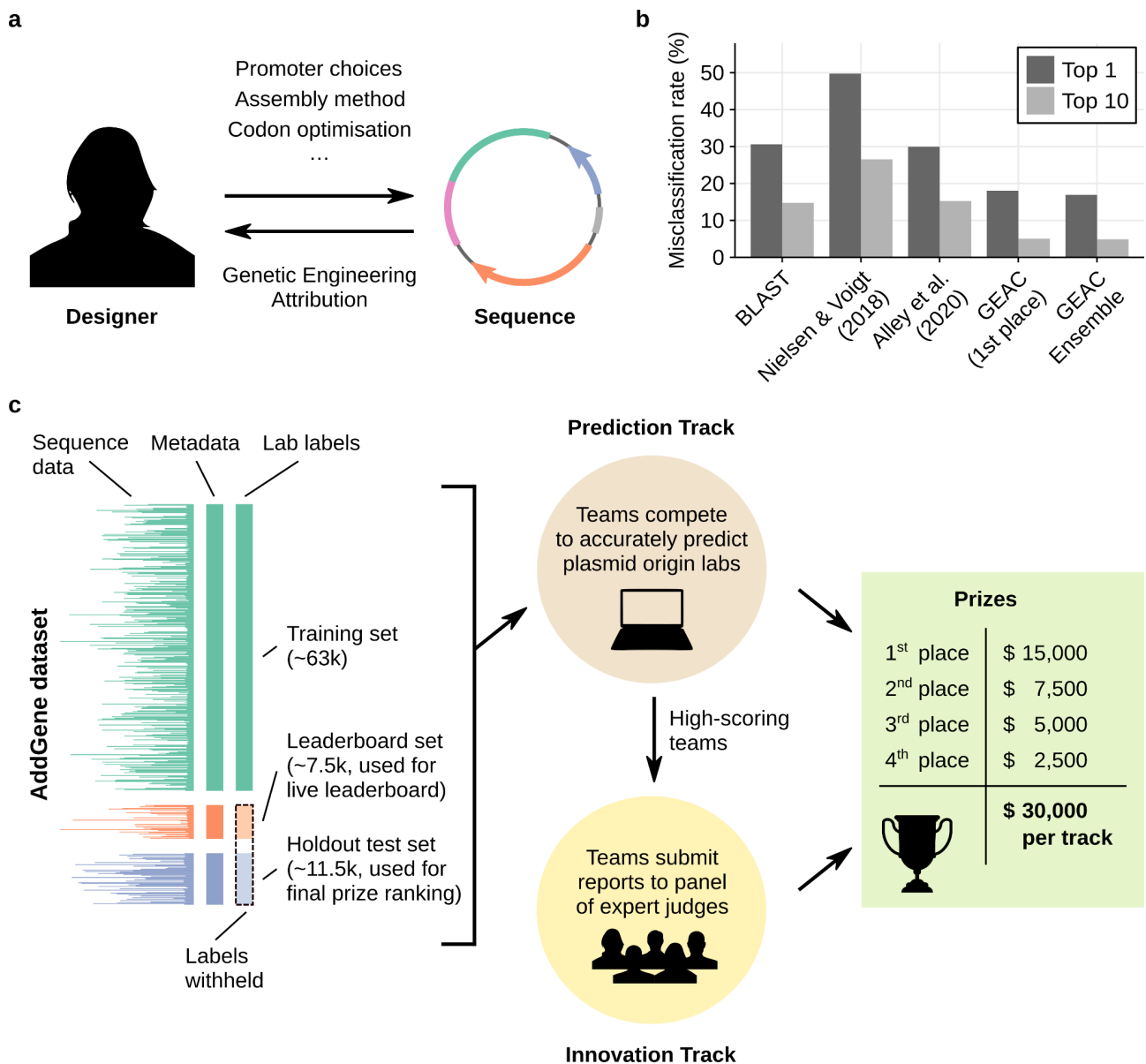

**Figure 1: The Genetic Engineering Attribution Challenge.** (a) The creation of any synthetic nucleic-acid sequence involves numerous design decisions, each of which leaves a mark in the resulting sequence. Genetic engineering attribution (GEA) aims to use these marks to identify the designer. (b) Misclassification rate (1-(Top-N accuracy)) of past ML approaches to GEA on the Addgene plasmid database, compared to BLAST (left) and the results of the Genetic Engineering Attribution Challenge (GEAC, right). Lower misclassification rates indicate higher accuracy. Our BLAST method achieves higher accuracy than previous implementations; see Online Methods for details. (c) In the GEAC, teams were provided with engineered plasmid sequences from Addgene, alongside basic metadata for each plasmid. Lab-of-origin labels were provided for the training dataset, but withheld from the leaderboard and holdout test datasets. In the Prediction Track, teams competed to identify these withheld labs-of-origin with the greatest top-10 accuracy. In the Innovation Track, high-scoring teams from the Prediction Track were then invited to submit reports describing their approaches to a panel of expert judges for assessment.

A recent publication using a non-machine-learning (ML) pan-genome method reported comparable results, with 76% accuracy (henceforth, "top-1 accuracy") and 85% top-10 accuracy[6].

Inspired by these results and the success of past citizen science initiatives[7–10], we took a community-led approach to the problem, running the first Genetic Engineering Attribution Challenge (GEAC, Fig. 1c) in July-November 2020 (Online Methods). This public data-science competition, hosted on the DrivenData online platform[11], consisted of two sequential tracks, termed the Prediction Track and the Innovation Track. In the Prediction Track, teams competed to predict the lab-of-origin of plasmid sequences with the highest possible top-10 accuracy. High-scoring teams from the Prediction Track were then invited to participate in the Innovation Track, writing short reports on their approaches which were assessed by a multidisciplinary panel of expert judges. A prize pool of $30,000 was offered for each track (Supplementary Table 1).

We focus here on the results of the Prediction Track, which received more submissions and is more amenable to quantitative analysis. The dataset for the Prediction Track was derived from the Addgene dataset used by Alley et al.[2], comprising sequences and minimal metadata from 81,833 plasmids (Online Methods). These plasmids were deposited by 3,751 origin labs; labs with fewer than 10 plasmids were pooled into an auxiliary category (labelled "Unknown Engineered"), leaving a total of 1,314 categories for classification. The dataset was divided into training, leaderboard, and holdout test sets (Fig. 1c), with top-10 accuracy on the holdout set determining the final ranking.

## Results

**Core competition outcomes**

Over 1200 users, from 78 countries (Fig. 2a, Supplementary Table 2 & 3), registered to participate in the competition. Of these, 318 users, organised into 299 teams, made at least one submission. There was a strong positive correlation between the number of submissions made by a team and their final top-10 accuracy (Spearman's $\rho$ = 0.82, Fig. 2b, Supplementary Fig. 1): the mean number of submissions made by the top 10% of teams was 49.1, compared to 8.8 for the bottom 90% of teams and 1.4 for the bottom 10%.

The accuracies achieved by Prediction Track teams far exceeded previous work (Fig. 2c-d, Supplementary Fig. 2-3). 75 teams (25%) achieved higher top-10 accuracies than any previous ML-based GEA model[2,5]; the top-10 accuracy of the highest-scoring team (94.9%) exceeded the previous published record by over 10 percentage points. The other three prizewinning teams also achieved very high prediction accuracy, with top-10 accuracies ranging from 93.0% to 94.4% – all of which exceed the previous record by at least 8 percentage points.

While a single, simple scoring metric was required for the competition, top-10 accuracy represents only one perspective on the performance of an attribution model. To investigate whether the gains seen in this metric represent robust improvements in performance, we broadened our analysis to include top-N accuracy for different values of N (Fig. 2c-d, Supplementary Fig. 2-4). The best models from the competition outperformed previous work across a wide range of N-values – in the case of top-1 accuracy, for example, 40 teams (13.4%) outperformed the published record, with the top-scoring team's accuracy (81.9%) exceeding it by over 11 percentage points. A similar degree of outperformance was observed for top-5 and top-20 accuracy (Supplementary Fig. 3).

In addition to improved accuracy, the best models from the Prediction Track also outperformed previous work on other measures of model performance. The first-, second-, and fourth-place teams all exhibited higher precision and recall than the best previous model, and all four prizewinning teams outperformed the previous best F1 score (Fig. 2e-f, Supplementary Fig. 5-7). As with previous GEA models, most submissions exhibited higher precision than recall, indicating that they returned a higher rate of false negatives than false positives. This tendency can be counterbalanced by looking at a larger number of top predictions from each model – that is, by measuring top-N accuracy for N>1.

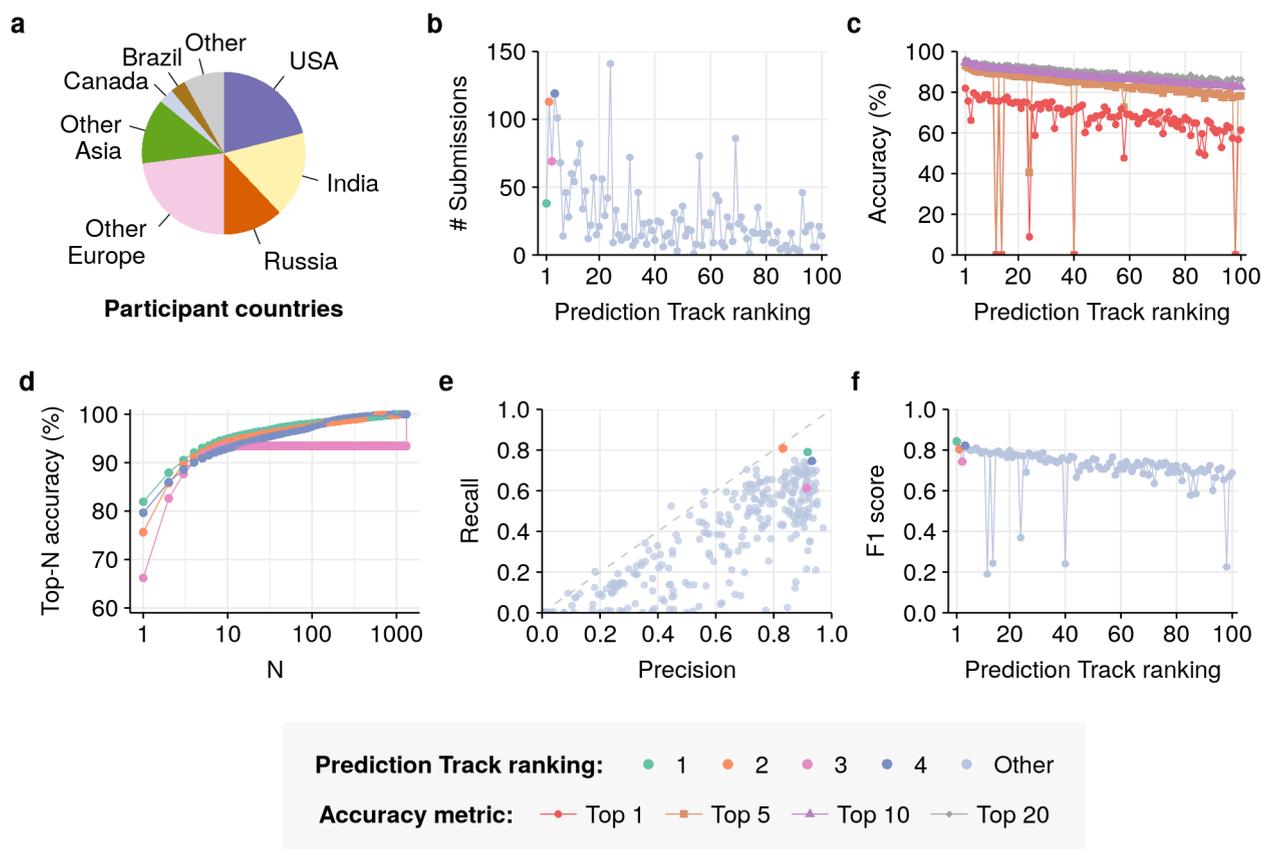

**Figure 2: Key Competition Results.** (a) Countries of residence of registered competition participants. (b) Total number of submissions made by the top 100 Prediction Track teams. (c) Top-1, -5, -10 and -20 accuracy achieved by each of the top 100 Prediction Track teams. Top-10 accuracy (purple) was used to determine overall ranking and prizes. (d) Top-N accuracy curves of the four prize-winning submissions to the Prediction Track, as a function of N (e) Precision and recall of all 299 Prediction Track teams. Dashed grey line indicates *x=y*. (f) Macro-averaged F1 score achieved by each of the top 100 Prediction Track teams.

**Evaluating negative attribution with rank metrics**

In many important practical applications of genetic engineering attribution, the ability to confidently *exclude* a potential designer (so-called "negative attribution") can be extremely valuable, even if the true designer cannot be identified with confidence[4]. In these contexts, a longer list of candidates presented with extremely high confidence may be more useful than a shorter list presented with lower confidence.

To investigate the degree to which Prediction Track models enable this sort of confident negative attribution, we developed a new metric. The *X99 score* of a predictor is the minimum positive integer N such that the top-N accuracy of that predictor is at least 99% (Fig. 3a). Analogous metrics can be defined for other accuracy thresholds; for example, the X95 score of a predictor is the smallest value of N such that its top-N accuracy is at least 95%. The lower the values of these two metrics, the better the predictor is able to confidently focus subsequent investigation on a manageable set of candidates.

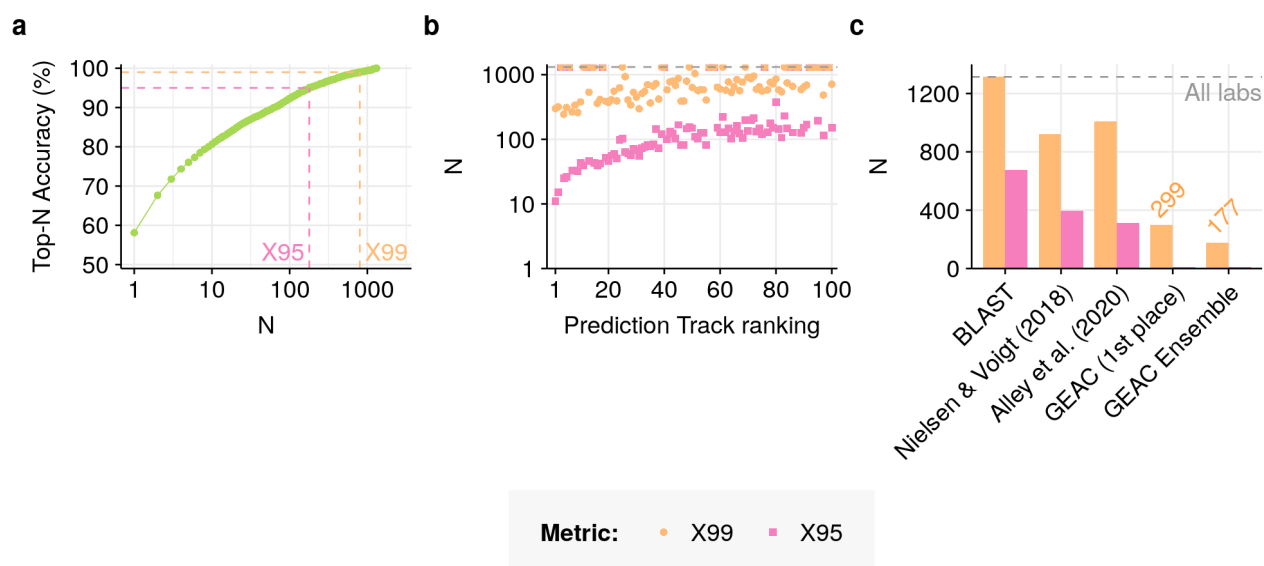

**Figure 3: Rank metrics for efficient genetic forensics.** (a) For any given lab-of-origin predictor, the X99 score is the smallest positive integer N such that the top-N accuracy of the predictor is at least 99%. Analogous metrics can be defined for other thresholds; for example, the X95 score is the smallest N such that top-N accuracy exceeds 95%. (b) X99 & X95 scores achieved by each of the top 100 Prediction Track teams, on a logarithmic scale. (c) X99 & X95 scores achieved by past ML-based approaches to GEA on the Addgene plasmid database, compared to BLAST (left) and the results of the Genetic Engineering Attribution Challenge (GEAC, right). X99 results for the GEAC 1st place and ensemble models are annotated in orange. Dashed grey horizontal line in (b-c) indicates the total number of labs in the dataset, which represents the largest possible value of any X-metric on this dataset.

We computed X99 and X95 scores for every team in the Prediction Track, as well as for previously published GEA models (Fig. 3b-c, Fig. 4, Supplementary Fig. 8-13). The lowest X99 score achieved by any previous model on the same dataset was 898 (using the CNN model of Nielsen & Voigt 2018), while the lowest previous X95 score was 311 (using the RNN model of Alley et al. 2020). In contrast, the lowest X99 score achieved in the Prediction Track was 244, achieved by the fourth-place Prediction Track team – a 73% reduction compared to the previous record. The X99 score of the first-place team was 299. The lowest X95 score achieved in the Prediction was 11, achieved by the first-place team – a 96% reduction. The competition results thus represent a dramatic improvement in negative attribution capability.

**Improving performance with ensembling**

Ensembles of multiple models routinely improve performance across a wide range of machine-learning problems[12–14]. Indeed, all prizewinning teams in the Prediction Track made use of some sort of ensemble to generate their predictions (see below). We therefore hypothesised that further ensembling could achieve even greater performance.

Our simple ensemble of the winning models (Online Methods) achieved marginally higher top-10 accuracy than the 1st-place team, showing a gain of 0.2 percentage points (95.1 vs 94.9%, Fig. 1b, Supplementary Fig. 3 & 14). The improvement seen in top-1 accuracy was larger, with an increase of 1.4 percentage points (83.1% vs 81.9%). This degree of top-1 accuracy approaches the best top-10 accuracies previously reported in the literature[2,6]. The ensemble model also achieved the highest F1 score of any ML-based GEA model to date (Supplementary Fig. 6), reflecting a better balance between precision and recall than was achieved by individual winning teams.

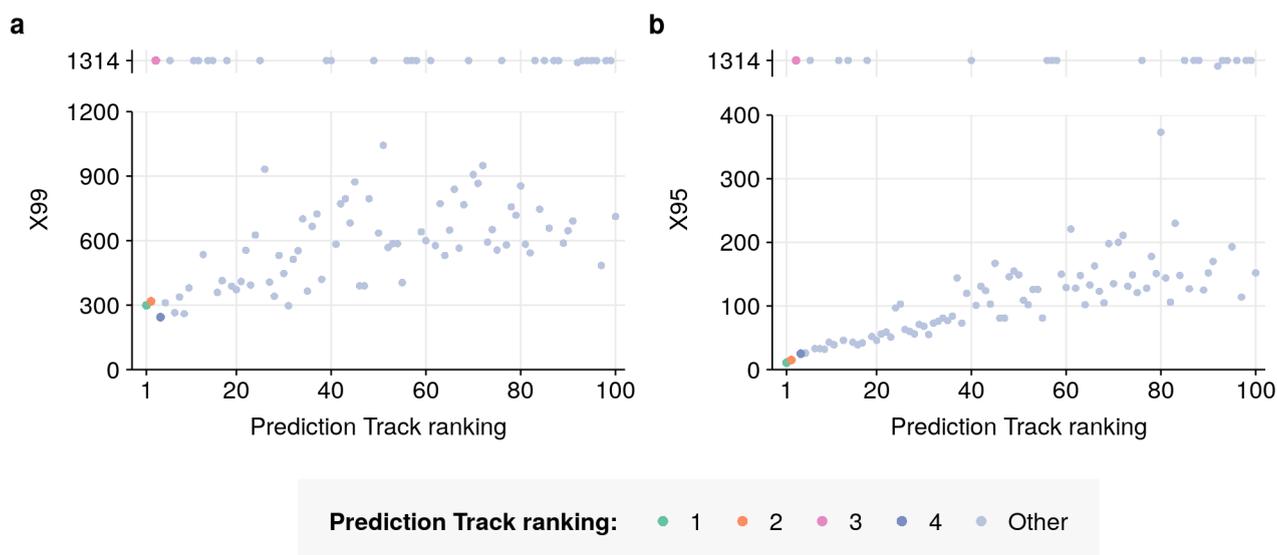

**Figure 4. X-metrics in detail.** (a) X99 and (b) X95 scores achieved by each of the top 100 Prediction Track teams, on separate linear scales. In each case, outlier values are shown in a separate panel above. There are 1,314 lab categories in the dataset.

By far the largest improvement from the ensemble was seen in the X99 negative-attribution metric discussed above (Fig. 3c, Supplementary Fig. 9). The ensemble achieved an X99 score of 177, compared to 299 for the overall competition winner and 244 for the team with the lowest X99 (a 27.5% reduction). This dramatic improvement suggests that significant further gains in X99 may be possible, further increasing the practical applicability of GEA models.

**Effects of large composite classes on prediction accuracy**

As discussed above, small labs in the competition dataset were pooled into a single auxiliary category, labelled "Unknown Engineered". This category was the largest in the dataset, making up 7.5% of sequences, compared to 2.4% for the largest unique lab (Supplementary Fig. 15). Given this frequency, it is possible that teams could inflate their Prediction Track scores by always including Unknown Engineered in their top 10 lab-of-origin guesses. Indeed, high-scoring teams included Unknown Engineered in their top-10 guesses at a rate far exceeding its true frequency, and the frequency with which they did so was correlated with their overall top-10 accuracy (Spearman's $\rho$ = 0.57, Supplementary Fig. 16a-b).

As a result, the top-10 accuracy achieved by most teams on Unknown Engineered sequences far exceeded that of sequences assigned to a unique lab category, inflating teams' top-10 accuracy overall (Supplementary Fig. 16c). Previous GEA models exhibited similar behaviour (Supplementary Fig. 17). In general, however, the effect was marginal: for the top 10% of teams, the average top-10 accuracy on unique (non-Unknown-Engineered) labs was only 0.7 percentage points lower than their accuracy on the entire dataset. Nevertheless, these results illustrate an important weakness in this approach to handling small and unseen labs in GEA datasets.

**Calibration of competition models**

Deep-learning models are often overconfident in their predictions[15]. This can cause problems for their interpretation, especially in cases, like GEA, where the evidence from such models needs to be weighed alongside multiple other data sources. Under these circumstances, it is useful to measure the calibration of model predictions, and potentially to take steps to improve that calibration prior to use[15–17].

Under conventional definitions of calibration, a predictor is considered to be well-calibrated if events it predicts with probability $Y$ occur $100 \times Y$% of the time. Common metrics for measuring calibration in this vein include the Expected Calibration Error (ECE) and Maximum Calibration Error (MCE)[15], which measure the average and maximum absolute deviation observed across some number of binned ranges (Online Methods).

Previous work on genetic engineering attribution has included calibration analysis. Alley et al.[2] found that their RNN-based model was reasonably well-calibrated (ECE = 4.7%, MCE = 8.9%); our reanalysis of that model's predictions returned similar values (ECE = 5.9%, MCE = 8.9%, Supplementary Fig. 18). We also found that this RNN model was far better calibrated than other previous attempts at GEA, especially with regard to MCE (Supplementary Fig. 18). Given these results, we decided to investigate the calibration of Prediction Track teams.

The MCEs and ECEs exhibited by Prediction Track teams varied widely, and were only weakly-to-moderately correlated with Prediction Track ranking (Spearman's $\rho$ vs ECE = 0.15, $\rho$ vs MCE = 0.38, Supplementary Fig. 19). Among the prizewinning teams, the 4th-place winner performed best in terms of calibration, achieving results comparable to Alley et al. (ECE = 3.4% and MCE = 11.8%, Supplementary Fig. 18). The other prizewinners exhibited worse performance, with an average ECE of 23.5% and an average MCE of 27.7%. This reflects generally poor calibration among teams generally: the top 10% of teams achieved an average ECE of 17.5% and an average MCE of 33.5% (Supplementary Fig. 20).

These results are not surprising: it is common for deep-learning models to be very miscalibrated[15], and models in the Prediction Track were not penalised for poor calibration. Nevertheless, our results demonstrate that the relative rankings produced by these models are generally more informative than their specific probability estimates.

**Strategies used by prize-winning teams**

At the close of the competition, the prizewinning teams shared their model code with organisers, allowing us to investigate the strategies they employed[18]. At a high level, the 1st-, 2nd- and 4th-place teams took remarkably similar approaches, with all of them employing ensembles containing at least one convolutional network[12,13,19]. However, the precise structure of these ensembles, including the number and size of the component networks[14] and the preprocessing methods employed, varied considerably. Several teams normalised or augmented their dataset using the reverse complement of each sequence, and one team used principal component analysis[20] on BLAST features as input to their neural network. The 1st-place team combined multiple CNNs with a model based on $k$-mer counts, which appeared to complement the CNNs. Unlike the previous best-performing GEA model[2], none of the winning teams employed an RNN-based approach.

In sharp contrast to other winning teams, the 3rd-place Prediction Track team did not employ neural networks at all. Instead, they took a radically different approach, using $k$-mer kernels, naive Bayes[20], soft masks and rank merging[21]. In addition to achieving top-10 accuracy comparable with the best neural-network-based solutions, this approach was also dramatically faster: 0.66 CPU hours to train and run, compared to >40 GPU hours for similarly performant deep-learning-based solutions – a 1000-fold difference in the cost of compute (Online Methods). This approach had substantially worse top-1 accuracy (Fig. 2d) and X95/X99 scores (Fig. 4) than the other winning solutions; however, many of these may be the result of over-optimisation for the top-10-accuracy metric used in the competition, rather than inherent limitations.

## Discussion

By most quantitative metrics we investigated, the first Genetic Engineering Attribution Challenge was a resounding success. Along its core evaluation metric, top-10 accuracy, winning teams achieved dramatically better results than any previous attempt at genetic engineering attribution, with the top-scoring team and all-winners ensemble both beating the previous state-of-the-art by over 10 percentage points. Similarly large gains were seen for the more-conventional top-1-accuracy metric, despite submissions receiving no additional benefit from placing the true lab in first place.

To investigate whether models at this level of performance might be useful in practice, we developed two new metrics: X95 and X99. These metrics evaluate whether a model can generate a manageable list of candidates while reliably (with 95 or 99% confidence) including the true lab-of-origin. At the 95% level, the best models from the competition essentially solved this problem for the Addgene dataset, reducing X95 from over 300 to less than 15. Progress on X99 was similarly dramatic: our ensemble of the winning models achieved an X99 score of 177, an 80% reduction compared to previous work. Nevertheless, at the 99% level, further progress is needed before the problem can be considered solved.

While high-scoring competition teams performed extremely well on accuracy and X95/X99 metrics, not all the metrics we investigated showed such positive results. In particular, winning models were much less well-calibrated than some previously published models, making it difficult to take the specific probabilities of their predictions at face value. Recall and F1 scores also showed further room for improvement. These suboptimal results are not surprising: ECE, MCE, recall, and F1 all focus on the single top prediction made by a model for each sequence, but models in the competition were rewarded for ranking the true lab anywhere in their top 10 predictions. Future models, trained under broader optimisation incentives, will hopefully achieve similar or greater accuracy while excelling along a wider variety of metrics; further focus on X99 in particular could help reward models that are more robustly useful.

While the results of this competition are highly encouraging, it is important to keep in mind the gulf between the form of attribution problem presented here, and the problems to which GEA might be applied in practice. In many respects, the Addgene dataset – a large, well-curated database of broadly similar plasmid sequences, with the authorship of each sequence made freely available – represents a highly simplified form of genetic engineering attribution. While the availability of this dataset has been critical to the development of genetic engineering attribution approaches to date, if they are to be practically useful, attribution models will eventually need to generalise far beyond this initial scenario.

From this perspective of practical application, the fact that so many teams outperformed the previous best models in this field is promising, as it suggests that a wide variety of approaches could perform well on this problem. That one of the prizewinning teams adopted a very fast and completely neural-network-free approach to the problem is also encouraging, since speed of deployment and ease of retraining will be important in many applications of attribution technology. Future exploration of these and other desirable properties, alongside improvements in accuracy, will be an important part of bringing GEA into regular use.

At the same time, we envisage that investigating a wider range of methods, such as equivariant neural networks[22], transformers/attention methods[23] and uncertainty-aware approaches[24–27] may prove fruitful. Alternative approaches to handling small and unseen classes in GEA datasets – such as data augmentation[28,29], anomaly detection[30–32], or the use of more robust evaluation metrics[33,34] – should also be explored. Given the rapid improvement in genetic engineering attribution models to date, and the gains made during this competition, we are optimistic that further dramatic improvements, even to the point of practical application, may be within sight.

**Author Contributions:** ECA conceived the competition project & secured funding. ECA, JLS & WJB designed the competition with assistance from KLW, SRH, TGM & GL. WJB, TGB & CC drafted competition materials with assistance from other authors, especially ECA, JLS & SRH. WJB managed the competition project with assistance from KLW, ECA, JLS, SRH, TGM, & GL. DrivenData, via GL & CC, hosted the competition, recruited & managed participants, collected submissions, verified prizewinning submissions & disbursed prize money. OMC carried out analysis of competition data, with assistance from WJB. OMC and WJB generated figures. OMC and WJB wrote the manuscript, with input from all authors.

**Funding:** All authors were supported for their work on this project by Open Philanthropy. In addition, OMC was supported by a Todd-Bird Junior Research Fellowship from New College, Oxford, and CWB by a National Defense Science and Engineering Graduate Fellowship. Funding for competition prizes was provided by Open Philanthropy.

**Acknowledgements:** Plasmid data used in the competition were generously provided by Addgene, with Jason Niehaus providing invaluable data support. Grigory Khimulya & Michael Montague provided inspiration and advice throughout the project. Piers Millett, Matthew Watson, Thomas Inglesby, Claire-Marie Filone, Gregory Lewis, Gregory Koblentz, and David Relman all gave thoughtful advice on competition design and execution, while George M. Church and Nancy Connell gave thoughtful advice on competition related activities. Innovation Track judges, including several individuals mentioned above as well as Rosie Campbell, James Diggans, Gigi Gronvall, Jonathan Huggins, Joanna Lipinski, Natalie Ma, Adam Marblestone, and Carroll Wainwright, read and assessed submissions and provided thoughtful commentary. Finally, our thanks to all the participants in the Genetic Engineering Attribution Challenge, without whose engagement and creativity there would have been no competition.

## Online Methods

**Competition design**

*Overview*

The Genetic Engineering Attribution Challenge (GEAC) was a free, online, public data-science competition held on the DrivenData competition platform[11]. The competition was organised and sponsored by altLabs, Inc in collaboration with DrivenData, Inc. The competition was open to all individuals over the age of 18, from any country, with the exception of (i) officers, directors, employees and advisory board members of altLabs or DrivenData, (ii) immediate family members and housemates of those individuals, and (iii) individuals who are residents of countries designated by the United States Treasury's Office of Foreign Assets Control.

As discussed in the main text, the competition consisted of two sequential tracks: the Prediction Track and the Innovation Track, each of which is described in detail below. The Prediction Track ran from August 18 to October 19, 2020, while the Innovation Track ran from October 20 to November 1, 2020. Results for both tracks were announced on January 26, 2021. Both tracks had a total prize pool of US$30,000; the distribution of prize money among winning teams is specified in 1. All prize money was provided by altLabs, Inc.

*The Prediction Track*

In the Prediction Track, participants attempted to guess the lab-of-origin of plasmid sequences from the Alley et al. dataset (see below). Participants were given access to both training data and labels from the training set, while labels from the leaderboard and holdout test sets were withheld. The top-10 accuracy of each submission on the leaderboard set was immediately reported to the submitting team upon submission, and the best top-10 accuracy scores on this set for each team were continuously displayed on a public leaderboard during the competition. The top-10 accuracy of each submission on the holdout test set was not reported until after the Prediction Track had closed, and was used to determine the final competition ranking. Prizes were awarded to the four teams who achieved the highest top-10 accuracy scores on this private test set.

*The Innovation Track*

Following closure of the Prediction Track, teams that achieved a top-10 accuracy of at least 75.6% were invited to participate in the Innovation Track. This threshold was based on an earlier estimate of BLAST top-10 accuracy (see below). To compete in this track, participants were asked to submit short reports (maximum 4 pages, maximum 2 figures), which were then reviewed by a team of judges (see below). describing how their approach would contribute to solving real-world attribution problems. Prizes were awarded to teams who exhibited novel and creative approaches to the problem, or who demonstrated that their algorithms possessed useful properties other than raw accuracy. The full text of the Innovation Track problem description is available in the Supplementary Note.

Submitted reports were assessed by a team of twelve judges, including experts in synthetic biology, bioinformatics, biosecurity, and machine learning. Each judge reviewed a group of six submissions; assignment of submissions into these groups was performed randomly, with the constraints that each possible pair of submissions must be reviewed by at least two judges and that each individual submission must be reviewed by the same number of judges.

To avoid issues arising from differences in scoring practices between judges, each judge was asked to rank the submissions they received, with a rank of 1 indicating the best submission. Prizes were awarded to the four teams who achieved the smallest average rank across judges. In the event of a two-way tie, the process was repeated using only those judges who reviewed both submissions; this was sufficient to obtain four unique prizewinners in this case.

**Data preparation**

Data for the GEAC was provided by Alley et al.[2], and comprised all plasmids deposited in the Addgene repository up to July 27th 2018 – a total of 81,834 entries. For each plasmid, the dataset included a DNA sequence, along with metadata on growth strain, growth temperature, copy number, host species, bacterial resistance markers, and other selectable markers. Each of these categorical metadata fields was re-encoded as a series of one-hot feature groups:

- **Growth strain:** `growth_strain_ccdb_survival`, `growth_strain_dh10b`, `growth_strain_dh5alpha`, `growth_strain_neb_stable`, `growth_strain_other`, `growth_strain_stbl3`, `growth_strain_top10`, `growth_strain_xl1_blue`
- **Growth temperature:** `growth_temp_30`, `growth_temp_37`, `growth_temp_other`
- **Copy number:** `copy_number_high_copy`, `copy_number_low_copy`, `copy_number_unknown`
- **Host species:** `species_budding_yeast`, `species_fly`, `species_human`, `species_mouse`, `species_mustard_weed`, `species_nematode`, `species_other`, `species_rat`, `species_synthetic`, `species_zebrafish`
- **Bacterial resistance:** `bacterial_resistance_ampicillin`, `bacterial_resistance_chloramphenicol`, `bacterial_resistance_kanamycin`, `bacterial_resistance_other`, `bacterial_resistance_spectinomycin`
- **Other selectable markers:** `selectable_markers_blasticidin`, `selectable_markers_his3`, `selectable_markers_hygromycin`, `selectable_markers_leu2`, `selectable_markers_neomycin`, `selectable_markers_other`, `selectable_markers_puromycin`, `selectable_markers_trp1`, `selectable_markers_ura3`, `selectable_markers_zeocin`

In addition to the sequence and the above metadata fields, the raw dataset also contained unique sequence IDs, as well as separate IDs designating the origin lab. For the competition, both sequence and lab IDs were obfuscated through 1:1 replacement with random alphanumeric strings.

The number of plasmids deposited in the dataset by each lab was highly heterogeneous (Supplementary Figure 21). Many labs only deposited one or a few sequences – too few to adequately train a model to uniquely identify that lab. To deal with this problem, Alley et al. grouped labs with fewer than 10 data points into a single auxiliary category labelled "Unknown Engineered". This reduced the number of categories from 3751 (the number of labs) to 1314 (1313 unique labs + Unknown Engineered).

In addition to issues with small labs, the dataset also contains "lineages" of plasmids: sequences that were derived by modifying other sequences in the dataset. This could potentially bias accuracy measures by introducing dependencies between entries in the training and test sets. To deal with this issue, Alley et al. inferred lineage networks among plasmids in the dataset, based on information in the complete Addgene database acknowledging sequence contributions from other entries. More specifically, lineages were identified by searching for connected components within the network of entry-to-entry acknowledgements in the Addgene database (see Alley et al.[2] for more details).

The data were partitioned into train, validation, and test sets, with the constraints that (i) every category have at least three data points in the test set, and (ii) all plasmids in a given lineage be assigned to a single dataset. Following the split, the training set contained 63,017 entries (77.0%); the validation set contained 7,466 entries (9.1%); and the test set contained 11,351 entries (13.9%).

For the GEAC, these three data partitions were reassigned based on the needs of the competition: the training set was provided to the participants for model development, including the true (though obfuscated, see above) lab IDs. The validation and test sets, meanwhile, were repurposed as the leaderboard and holdout test sets of the competition. One entry with a 1nt sequence was dropped from the leaderboard set, leaving a total of 7,465 entries. The test and leaderboard sets were shuffled together, and provided to participants without the accompanying lab IDs; as described above, participants' top-10 accuracy on the validation set was used to determine their position in the public leaderboard during the competition, while their top-10 accuracy on the test set was used to determine the final ranking and prizewinners.

**Computing the BLAST benchmark**

Previous implementations of genetic engineering attribution using BLAST[35] have reported top-1 accuracies of just over 65% and top-10 accuracies of roughly 75%[2]. During the preparation of this manuscript, we found that a small modification of this attribution algorithm (specifically, replacing use of the quicksort algorithm[36] with mergesort[37]) resulted in equal top-1 accuracy, while substantially increasing top-N accuracy for N>1 (Supplementary Fig. 3). We have used the results from this modified algorithm in the main text, while presenting both sets of results side-by-side in the supplementary material. Under our implementation, the procedure followed by both algorithms can be summarised as follows:

- Sequences from the training set were extracted into a FASTA file, then used to generate a BLAST nucleotide database.
- Sequences from the test set were extracted into a FASTA file, then aligned to the training-set database, with an E-value threshold of 10.
- Alignments reported by BLAST were sorted in ascending order of E-value. The original implementation used quicksort for this sorting step, while our modified algorithm used mergesort. (In the latter but not former case, this is equivalent to sorting in descending order of bit score.)
- The lab IDs corresponding to each training-set sequence were identified, and the sorting results were filtered to include only the first result for each lab-ID/test-set-sequence combination. The remaining hits for each test-set sequence were ranked in ascending order of occurrence in the dataset.
- Finally, top-N accuracy was calculated as the proportion of test-set sequences for which the ID of the true origin lab was assigned a rank less than or equal to N.

For the purpose of calculating calibration (Supplementary Fig. 18), these ranks were reversed (so that the best match had the highest rank) and normalised using softmax.

**Other baselines**

Predictions on the competition test set for deteRNNt[2] and a reproduction of the CNN model developed by Nielsen & Voigt (2018)[5] were provided by Alley et al[2]. Top-N accuracy, X-metrics, calibration indices, and other metrics were re-computed from scratch based on these files.

**Post-competition analysis**

Each submission to the Prediction Track consisted of a *J* × *K* prediction matrix, where *J* is the number of sequences in the holdout test set (11,351) and *K* is the total number of lab classes in that test set (1,314). Each entry in this matrix ostensibly reflected a predicted probability of the corresponding lab being the true lab-of-origin for that sequence, with the entries in each row summing to unity.

To compute accuracy metrics for each team for this analysis, we first generated a rank matrix from their prediction matrix. In this matrix, the lab with the highest predicted probability for a given sequence was assigned rank 1, the second-highest prediction rank 2, and so on. To prevent teams achieving high scores by giving uniform predictions for large numbers of labs, tied predictions were assigned the maximum rank. Given this rank matrix, the top-N accuracy for any N could thus be computed as the proportion of rows for which the true lab was assigned a rank of N or less.

Given these accuracy scores, the X99 score could be computed as the minimum positive integer *N* such that top-*N* accuracy is at least 99%. This metric can be generalised to other thresholds, where X*R* is the minimum positive integer *N* such that top-*N* accuracy is at least *R*%. X95, X90 and X80 scores were all computed in this way.

For the purposes of calculating precision and recall, the number of true positives, false positives and false negatives were computed separately for each lab class for each submission. For a given class, the number of true positives $tp$ was defined as the number of times in the test set that that class was correctly assigned rank 1 (i.e. assigned rank 1 when it was in fact the true lab-of origin); the number of false positives $fp$ as the number of times it was incorrectly assigned rank 1, and the number of false negatives $fn$ as the number of times it was incorrectly assigned rank >1. Precision and recall for each class were then calculated as $tp/(tp+fp)$ and recall as $tp/(tp+fn)$, and the F1 score for each class as the harmonic mean of its precision and recall. The overall precision and recall for each team was computed as the arithmetic mean of its class-specific precisions and recalls, respectively, while the macro-averaged F1 score was computed as the arithmetic means of its class-specific F1 scores.

**Calibration**

Following Guo et al.[15] we checked whether predictions had frequentist calibration of their probabilistic forecasts. To estimate the expected accuracy from finite samples, we grouped predictions into 15 interval bins of equal size. We let $B_m$ be the set of indices of samples whose prediction confidence falls into the intervals $(m-1/M, m/M]$. The accuracy of bin $B_m$ is then defined as

$$\mathrm{acc}(B_m) = \frac{1}{|B_m|} \sum_{B_m} 1(\hat{y}_i = y_i),$$

where $\hat{y}_i$ and $y_i$ are the (top-1) predicted and true class labels for sequence $i$ and $|B_m|$ is the number of samples in bin $B_m$. The average confidence within bin $B_m$ is defined as

$$\mathrm{conf}(B_m) = \frac{1}{|B_m|} \sum_{B_m} \hat{p}_i,$$

where $\hat{p}_i$ is the predicted probability assigned to class $\hat{y}_i$ for sequence $i$. The expected deviation between confidence and accuracy can then be estimated using the expected calibration error (ECE):

$$\text{ECE} = \sum_{m=1}^{M} \frac{|B_m|}{n} |\text{acc}(B_m) - \text{conf}(B_m)|.$$

where $n$ is the total number of samples. The maximum calibration error (MCE) estimates the worst-case deviation from the binning procedure as:

$$\text{MCE} = \max_{m \in \{1,...,M\}} |\text{acc}(B_m) - \text{conf}(B_m)|.$$

**Ensemble**

To ensemble the four prizewinning teams from the Prediction Track, the probability assigned to each lab for each plasmid sequence was averaged between the top 4 classes with equal weight given to each class. That is the prediction for sequence $i$ to lab $j$ was given by:

$$p_{ij} = \frac{1}{4} \sum_{k=1}^{4} p_{ijk},$$

where $k$ indexes over the methods and $p_{ijk}$ is the prediction score given for sequence $i$ to lab $j$, by method $k$.

**Amazon web server compute costs**

Approximate costing for machine learning methods were calculated using Amazon EC2 on-demand pricing. We assumed a single GPU machine with sufficient memory (128 GB) costing $1.14 per hour (g3.8xlarge). This totals $51.30 for 45 hours of GPU time. For the CPU based methods, which required 20GB of solid-state drive (SSD) , an x2gd.medium instance, costing $0.08 per hour, would be sufficient. This totals $0.05 for the 0.66 CPU hours used.

**Data & code availability**

Summarised competition data & figure code are publicly available online at https://github.com/willbradshaw/geac/. Competition datasets and prediction data are available on request.

# Supplementary Tables

**Supplementary Table 1: Distribution of prizes.** The same distribution was used for both the Prediction Track and the Innovation Track. Both tracks had a total prize pool of $30,000, from an overall total of $60,000.

| Competition Rank | Prize money ($) |
|---|---|
| 1st | $15,000 |
| 2nd | $7,500 |
| 3rd | $5,000 |
| 4th | $2,500 |

**Supplementary Table 2: Competition engagement.** Number of users who viewed, joined, or submitted predictions to the Genetic Engineering Attribution Challenge. Many submitting teams consisted of a single individual, while some included multiple users.

| Activity | # Individuals |
|---|---|
| Visited competition website | 8101 |
| Registered as participant | 1211 |
| Submitted predictions (users) | 318 |
| Submitted predictions (teams) | 299 |

**Supplementary Table 3: Geographic origin of competition visitors and participants.** Proportion of site visitors and registered participants from each continent and country.

| Continent | Country / Region | % site visitors | % registered participants |
|---|---|---|---|
| North America | USA | 29 | 21 |
| | Canada | 3 | 3 |
| | Other N. America | < 1 | < 1 |
| South America | Brazil | 2 | 3 |
| | Other S. America | 2 | 1 |
| Asia | India | 13 | 17 |
| | Other Asia[1] | 15 | 13 |
| Europe | Russia | 8 | 12 |
| | UK | 4 | 3 |
| | Netherlands | 2 | 3 |
| | Germany | 3 | 3 |
| | France | 2 | 3 |
| | Other Europe | 12 | 11 |
| Africa | All countries | 3 | 5 |
| Oceania | All countries | 2 | 2 |

[1] China accounted for 2% of site visitors and ≤2% of participants.

**Supplementary Figures**

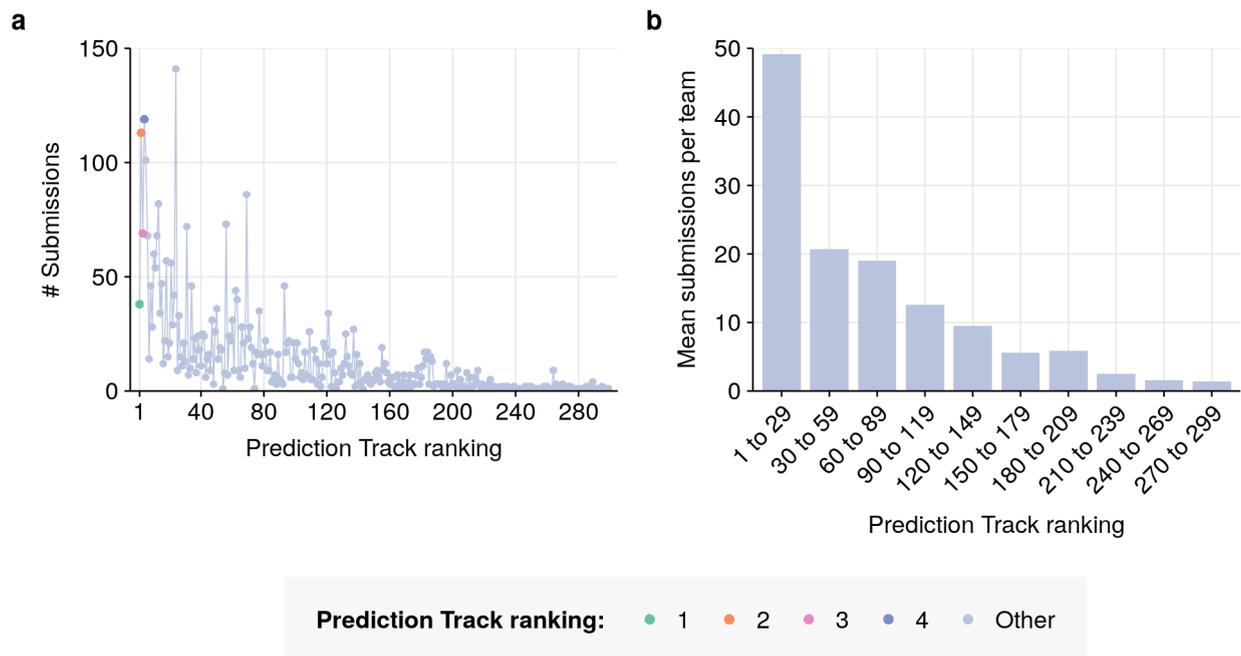

**Supplementary Figure 1: Submissions to the Prediction Track.** (a) Total number of submissions made by each Prediction Track team. (b) Mean number of submissions of each decile of teams, as ranked by top-10 accuracy.

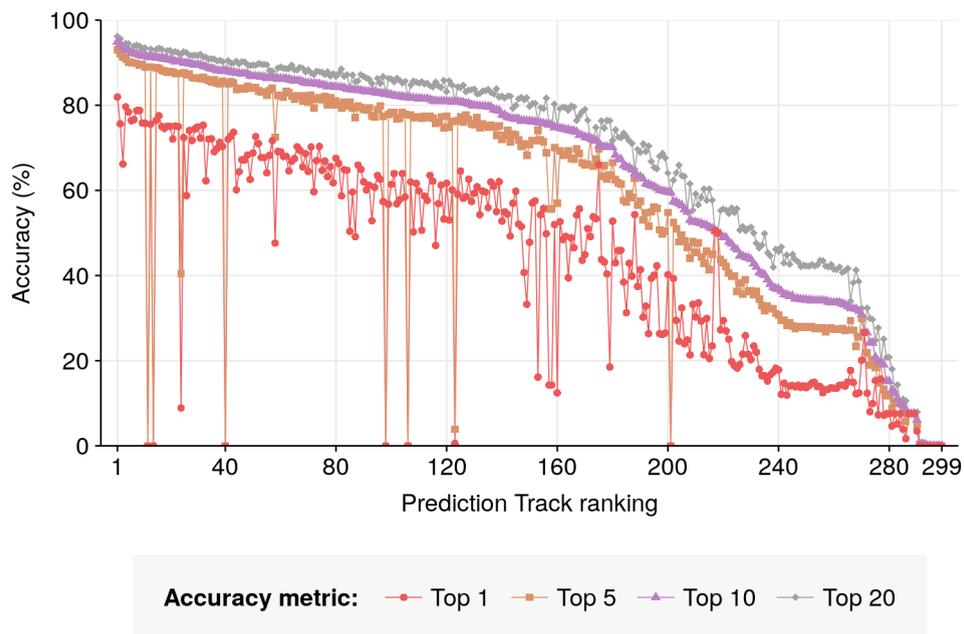

**Supplementary Figure 2: Prediction Track accuracy.** Top-1, -5, -10 and -20 accuracy achieved by each team in the Prediction Track.

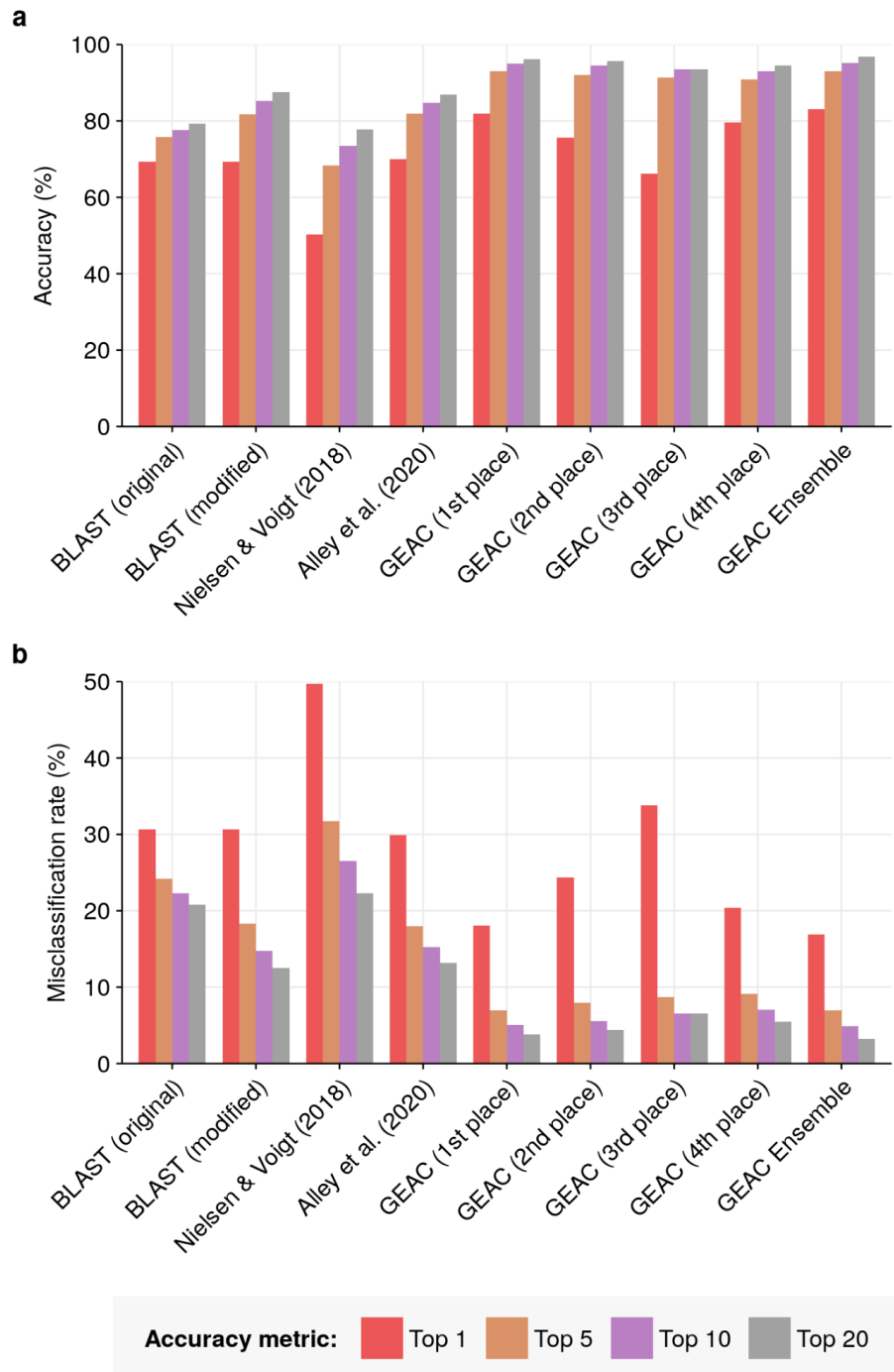

**Supplementary Figure 3: Key model accuracy.** (a) Top-1, -5, -10 and -20 accuracy achieved by Prediction Track winners and ensemble, as compared to BLAST (see Online Methods) and previously-published ML-based GEA models. (b) Misclassification rate (1-(Top-N accuracy)) for those models.

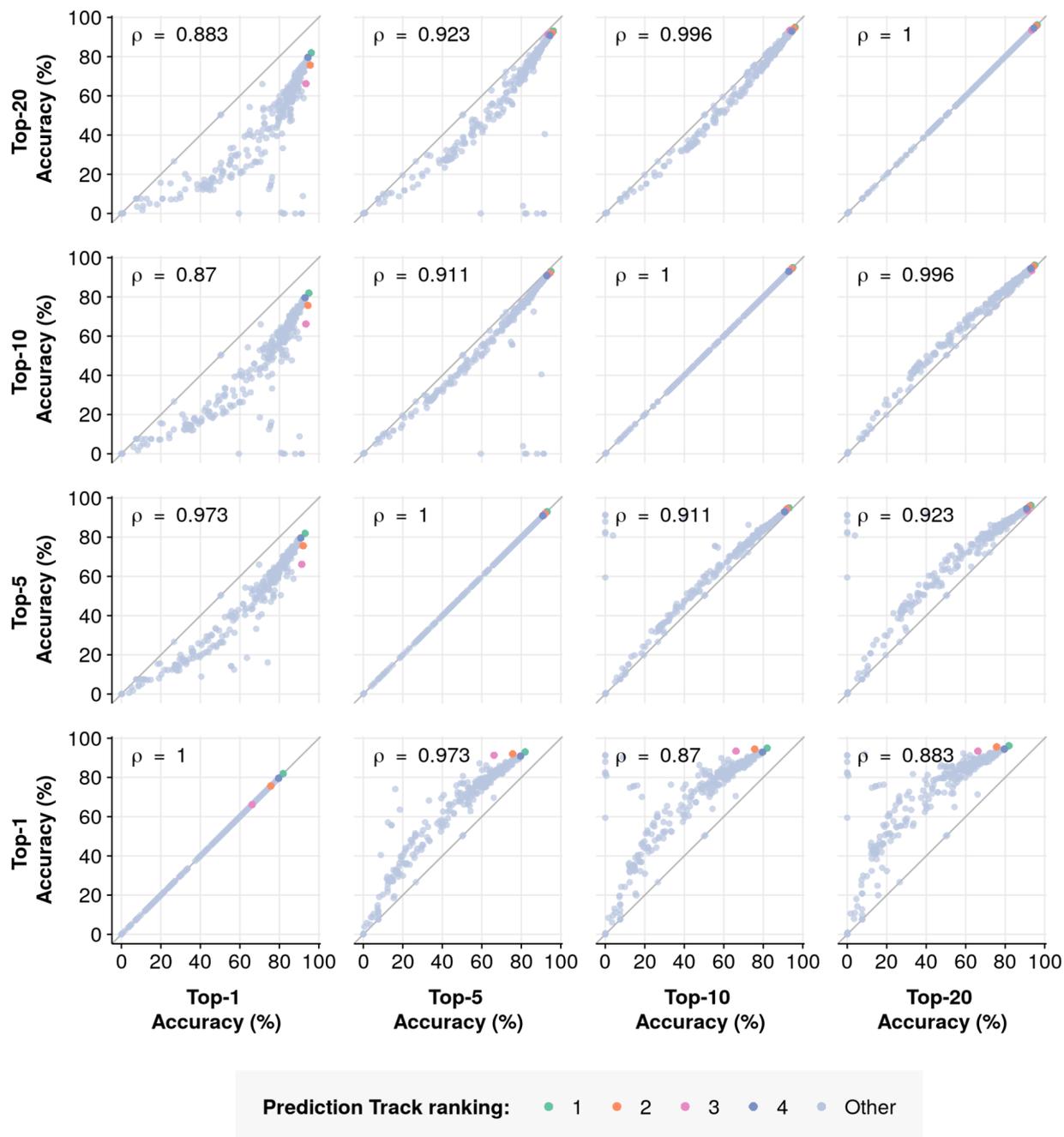

**Supplementary Figure 4: Relationship between Prediction Track accuracy metrics.** Scatter plots of the Top-1, -5, -10 and -20 accuracies achieved by each Prediction Track team, plotted against one another. Grey lines indicate *x=y*. Text in each panel indicates the Spearman rank correlation coefficient between each pair of metrics.

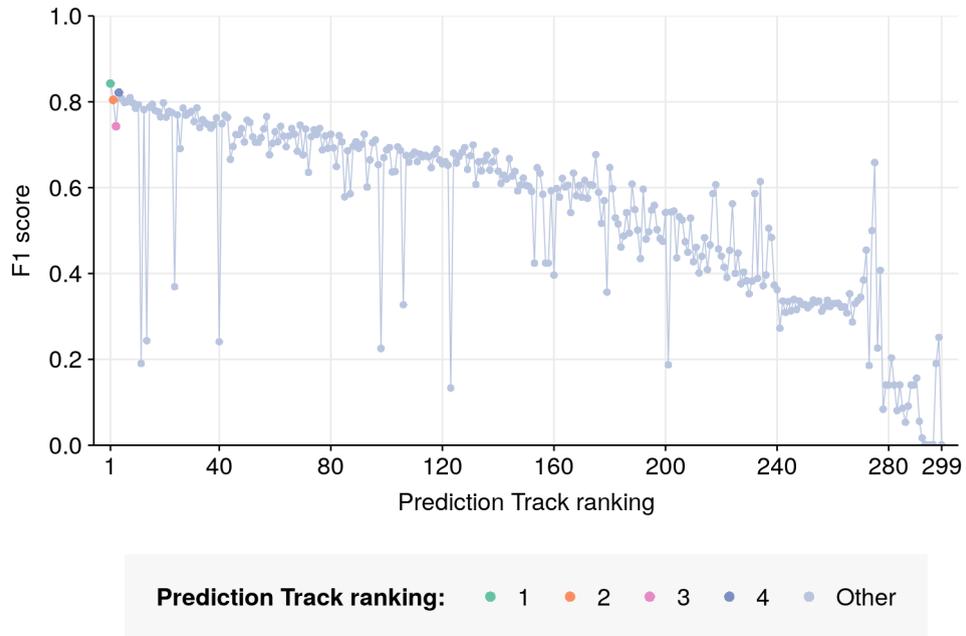

**Supplementary Figure 5: Prediction Track F1 scores.** Macro-averaged F1 score achieved by each team in the Prediction Track.

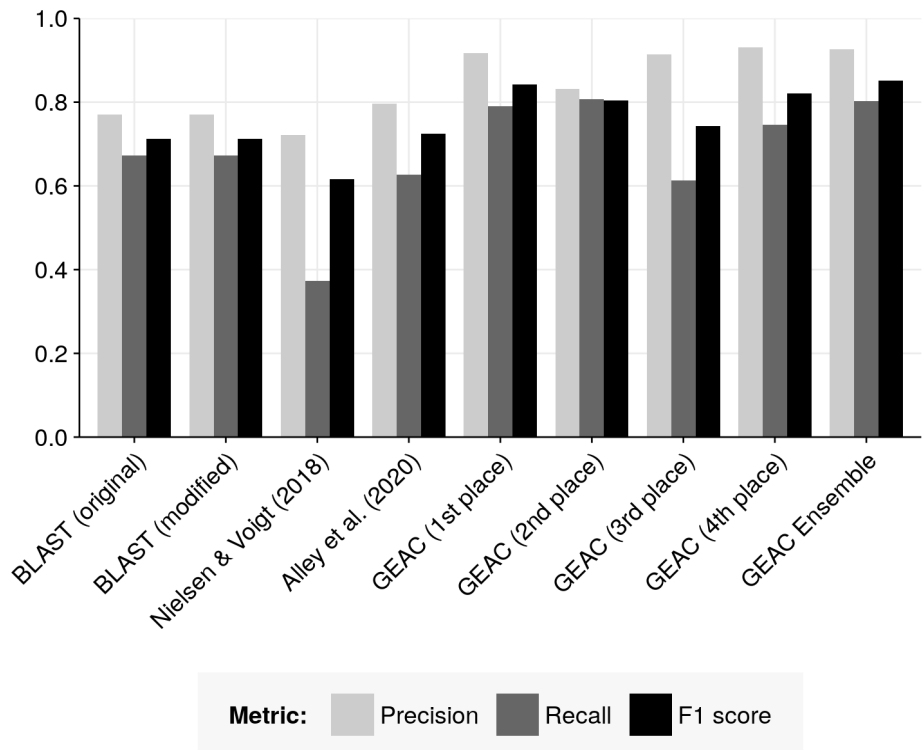

**Supplementary Figure 6: Precision, recall, and F1 for key models.** Precision, recall, and macro-averaged F1 scores achieved by Prediction Track winners and ensemble, as compared to BLAST and previously-published ML-based GEA models, on a logarithmic scale.

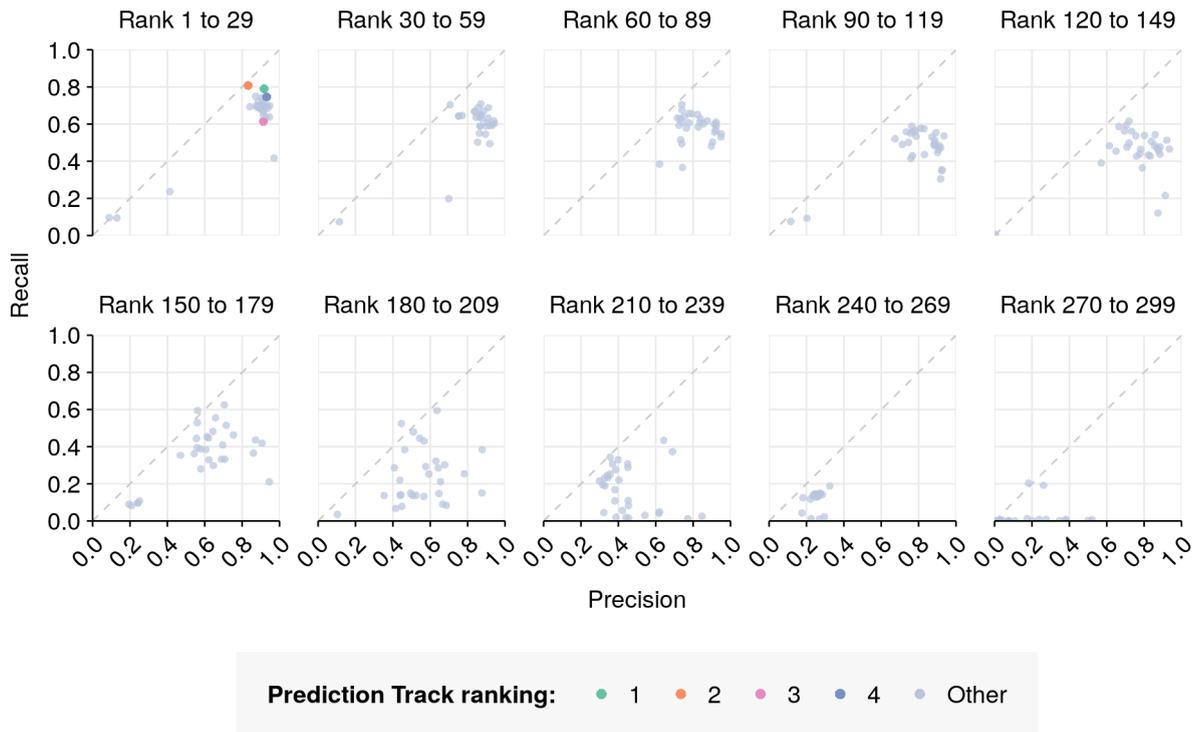

**Supplementary Figure 7: Prediction Track precision and recall.** Precision and recall achieved by each team in the Prediction Track, separated by score (i.e. top-10 accuracy) decile.

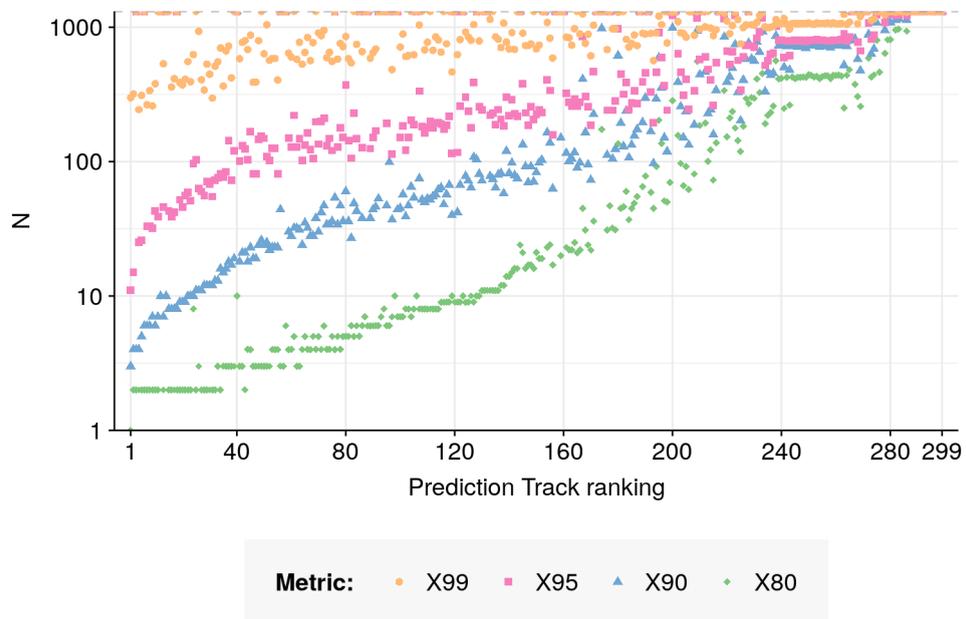

**Supplementary Figure 8: X-metrics for Prediction Track teams.** X99, X95, X90 and X80 scores achieved by each Prediction Track team, on a logarithmic scale.

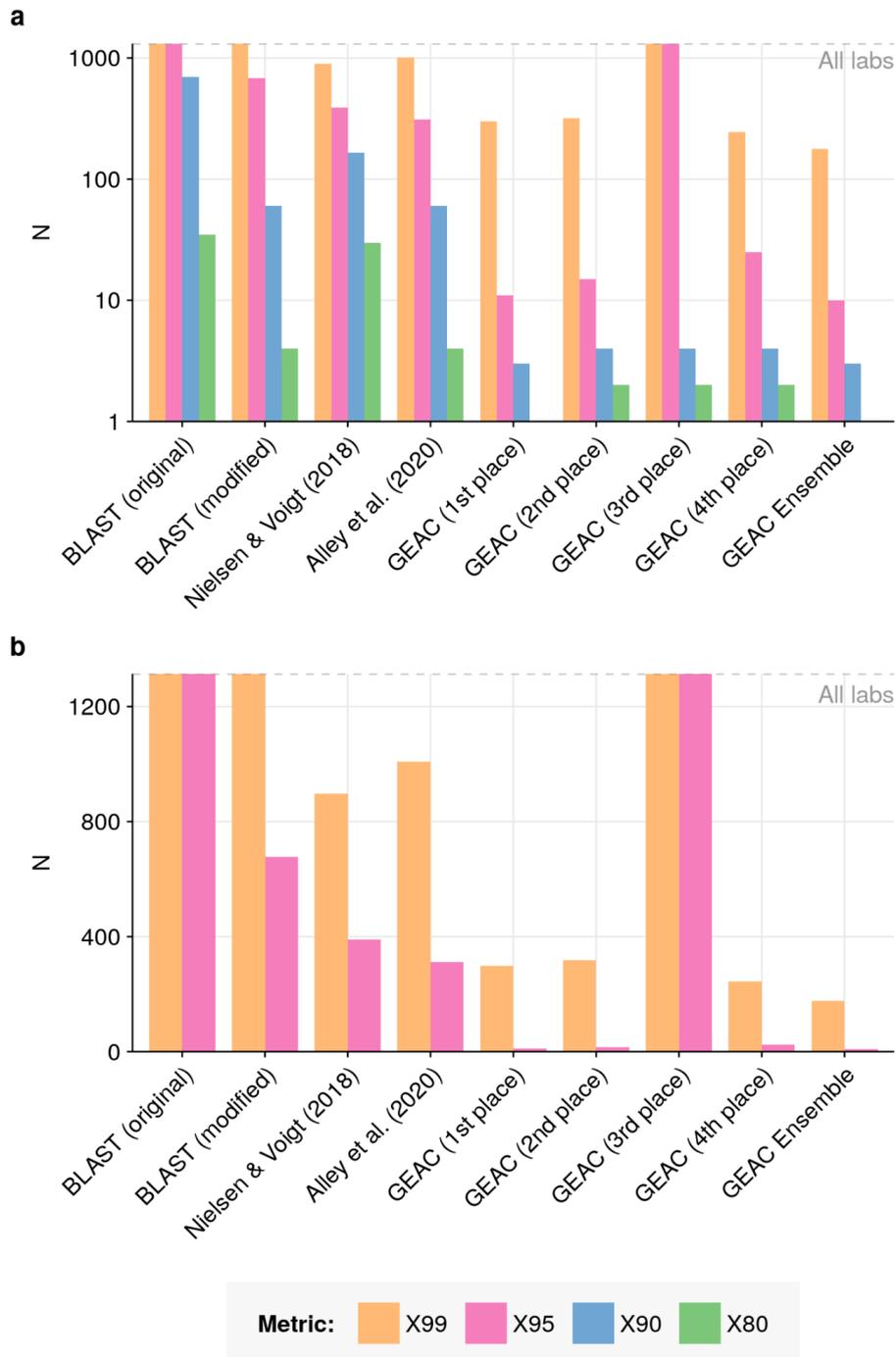

**Supplementary Figure 9: X-metrics for key models.** (a) X99, X95, X90 and X80 scores achieved by Prediction Track winners and ensemble, as compared to BLAST and previously-published ML-based GEA models, on a logarithmic scale. (b) X99 and X95 scores of those models, on a linear scale. Dashed grey horizontal lines indicate the total number of labs in the dataset, which represents the largest possible value of any X-metric on this dataset.

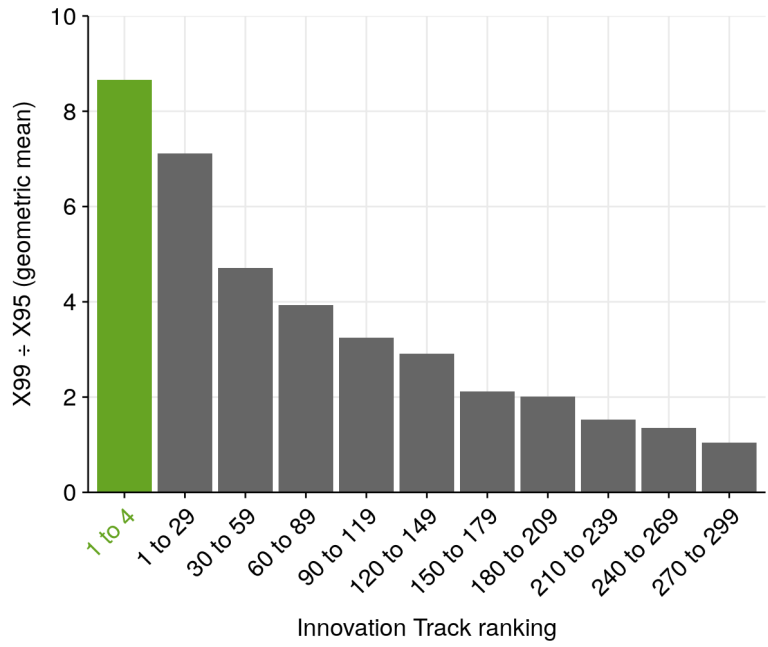

**Supplementary Figure 10: X99 vs X95.** Geometric mean of the ratio between X99 to X95 scores for teams in each rank decile of the Prediction Track (grey bars) as well as for the four winning teams (green bar).

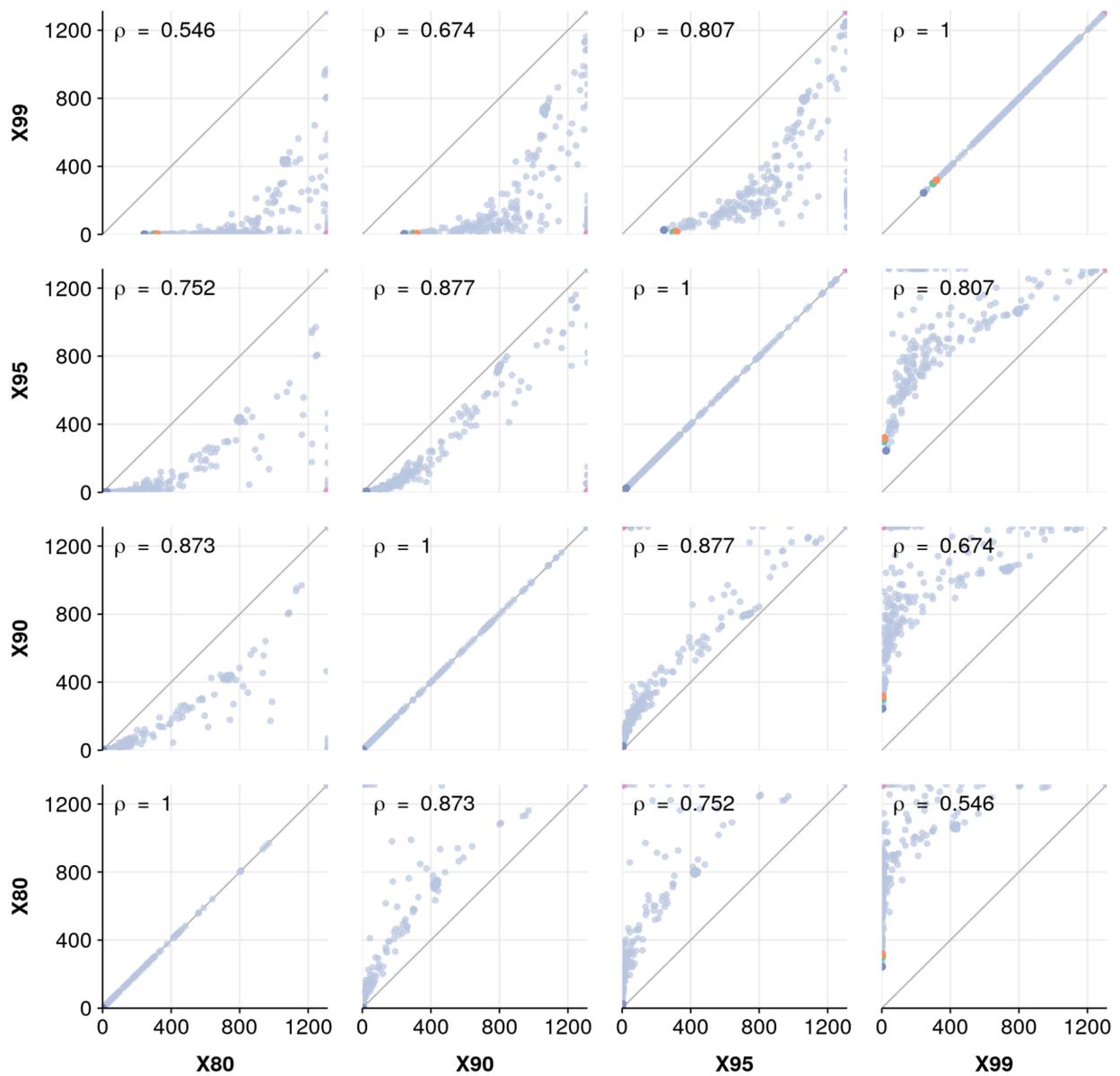

**Supplementary Figure 11: Relationship between Prediction Track X-metrics.** Scatter plots of the X99, X95, X90, and X80 scores achieved by each Prediction Track team, plotted against one another. Grey lines indicate *x=y*. Text in each panel indicates the Spearman rank correlation coefficient between each pair of metrics.

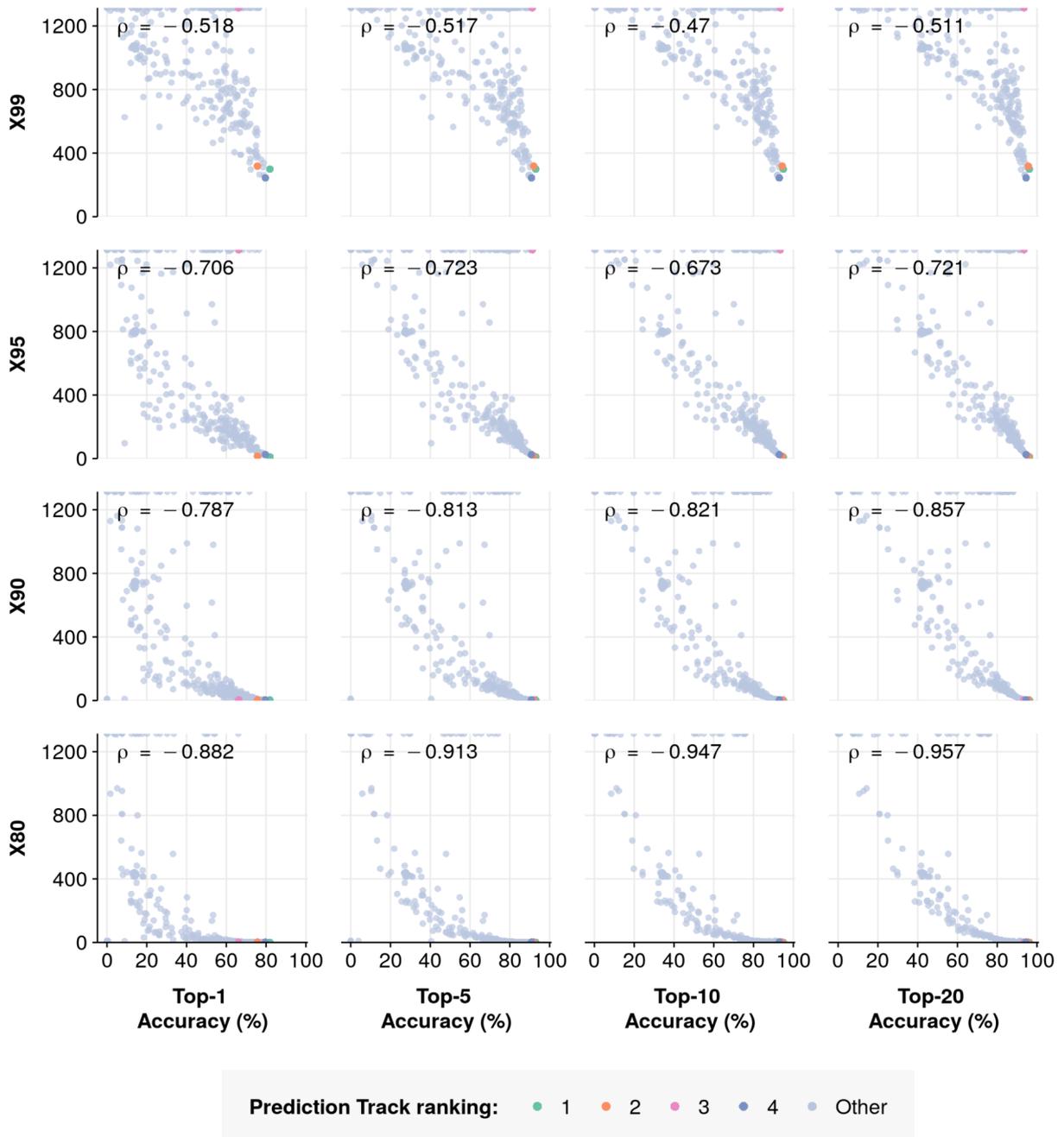

**Supplementary Figure 12: Relationship between Prediction Track accuracy and X-metrics (all teams).** Scatter plots of the X99, X95, X90, and X80 scores achieved by each Prediction Track team, plotted against their top-1, -5, -10 and -20 accuracies. Text in each panel indicates the Spearman rank correlation coefficient between each pair of metrics.

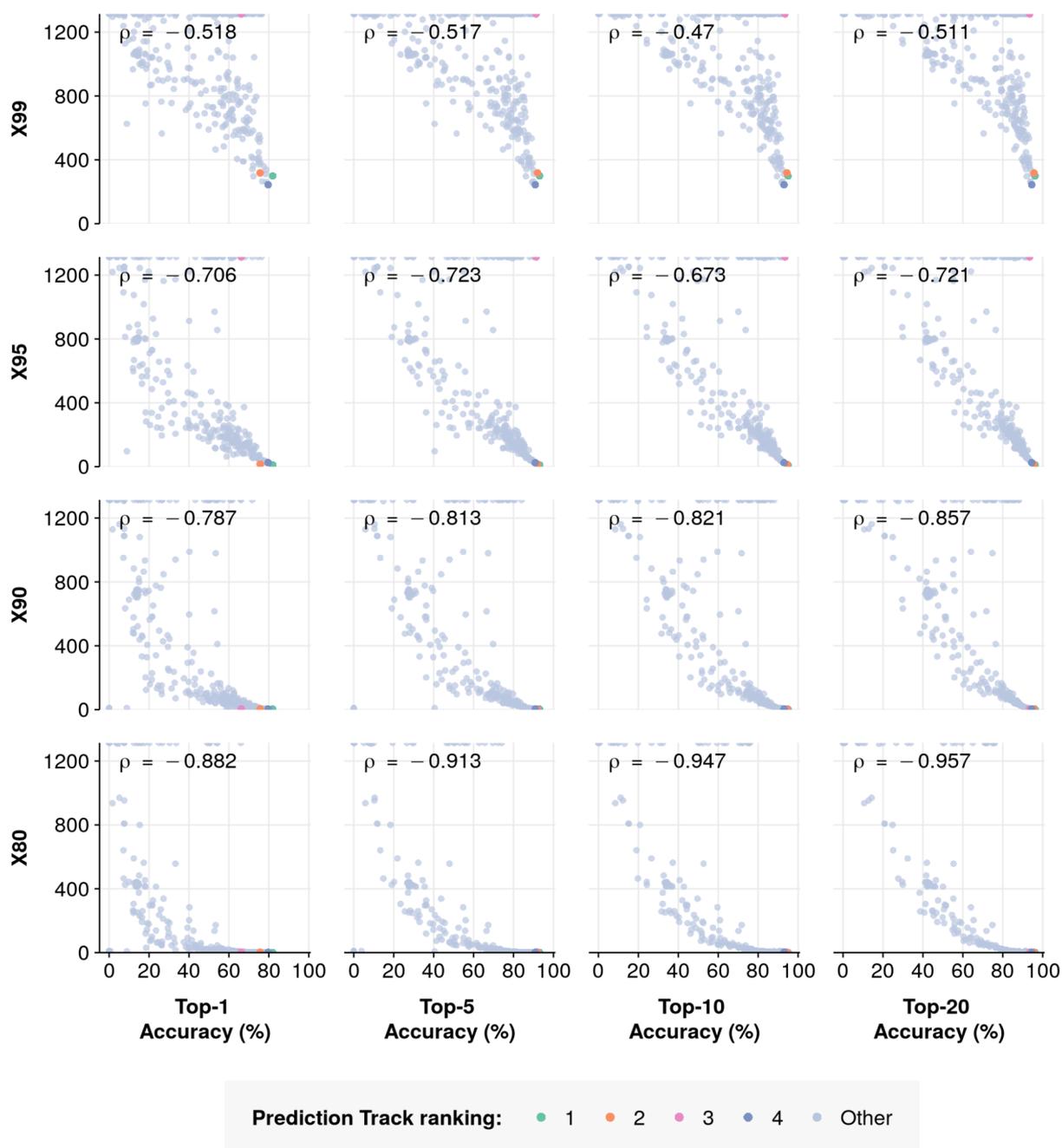

**Supplementary Figure 13: Relationship between Prediction Track accuracy and X-metrics (outliers removed).** Scatter plots of the X-scores and accuracy metrics achieved by each Prediction Track team, excluding those with X-metrics equal to the number of labs. Many teams only returned positive probabilities for their top 10 candidates for each sequence, preventing them from achieving 95% (or 99%) accuracy without including the entire set of labs as candidates. Text in each panel indicates the Spearman rank correlation coefficient between each pair of metrics, given the aforementioned filtering.

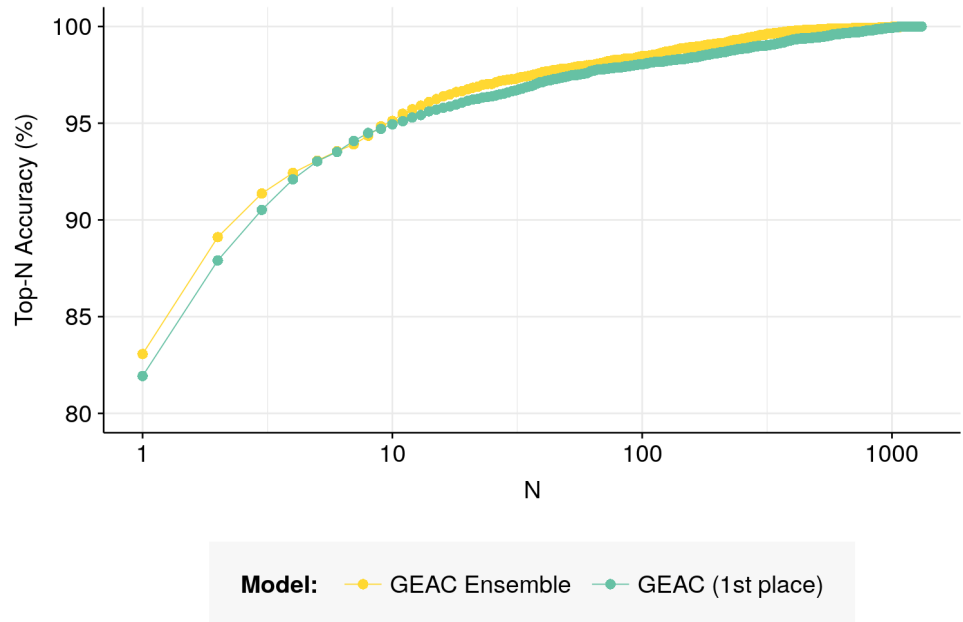

**Supplementary Figure 14: Ensemble top-N accuracy.** Top-N accuracy vs N for the GEAC ensemble model, as compared to the top-scoring Prediction Track team. For most values of N the Ensemble outperforms the competition winner.

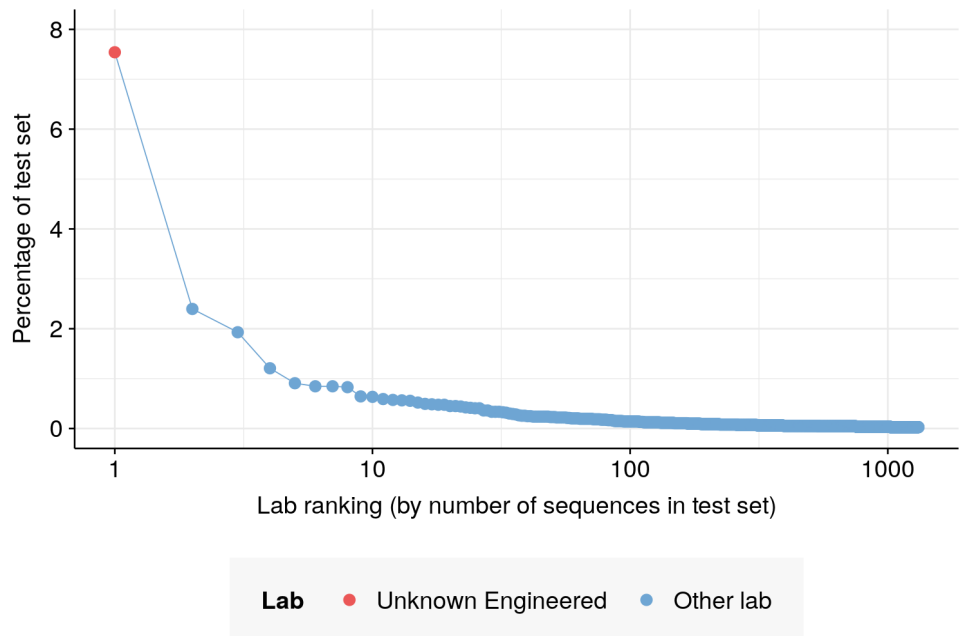

**Supplementary Figure 15: Lab composition of the GEAC test set.** Percentage of sequences in the competition test set accounted for by each lab category. Blue points indicate unique labs; the red point indicates the "Unknown Engineered" bucket category (Online Methods). Labs are ordered on the x-axis in descending order of prevalence in the test set.

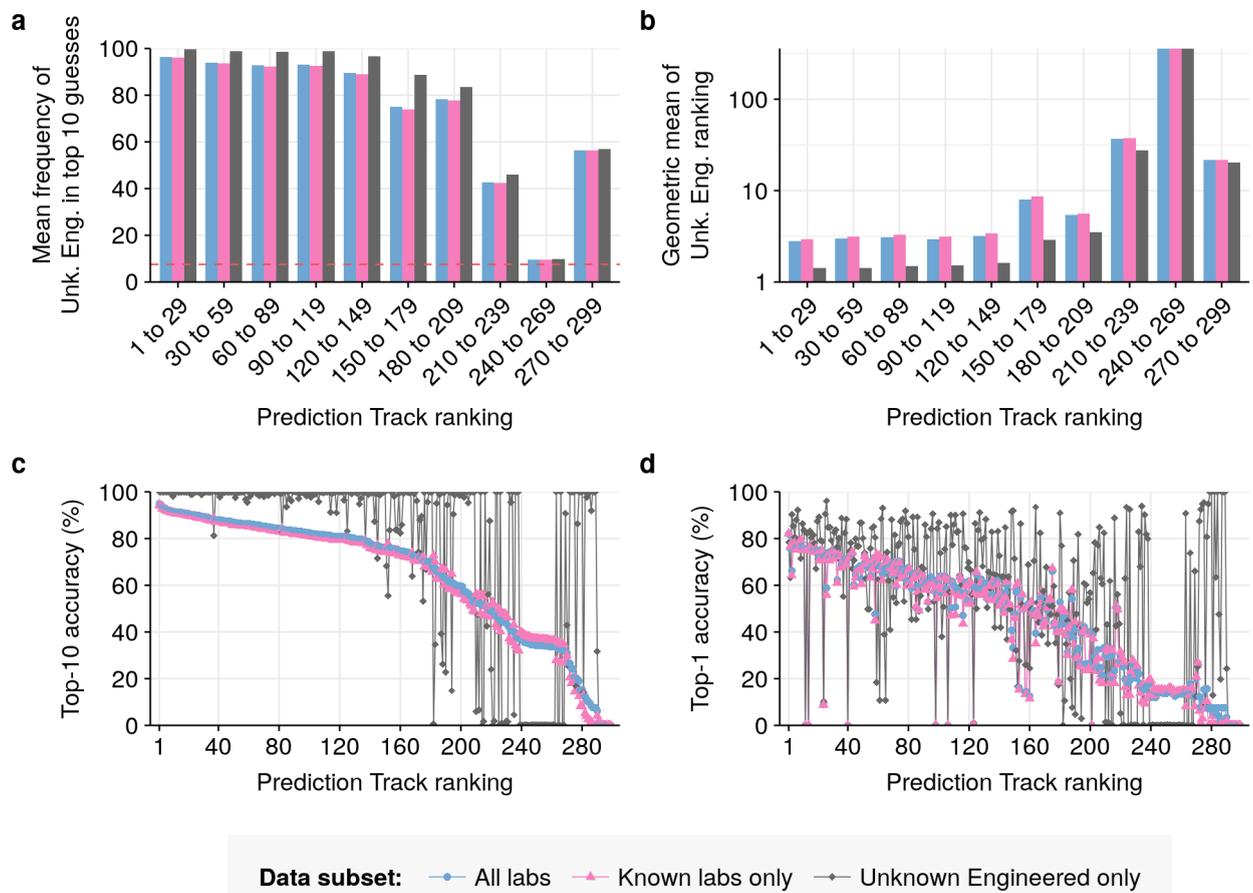

**Supplementary Figure 16: The Unknown Engineered category.** (a) Mean percentage of sequences in the test set for which submissions in each decile of teams included Unknown Engineered in their top-10 lab-of-origin guesses, compared to the true frequency of the category (7.5%, dashed red line). (b) Geometric mean rank of Unknown Engineered, across all sequences in the test set, for each decile of teams. (c) Top-10 accuracy achieved by each team in the Prediction Track on each of three subsets of the test set: sequences from all lab categories, sequences from known labs only (excluding Unknown Engineered), and Unknown Engineered sequences only. (d) Top-1 accuracies achieved on the same data subsets.

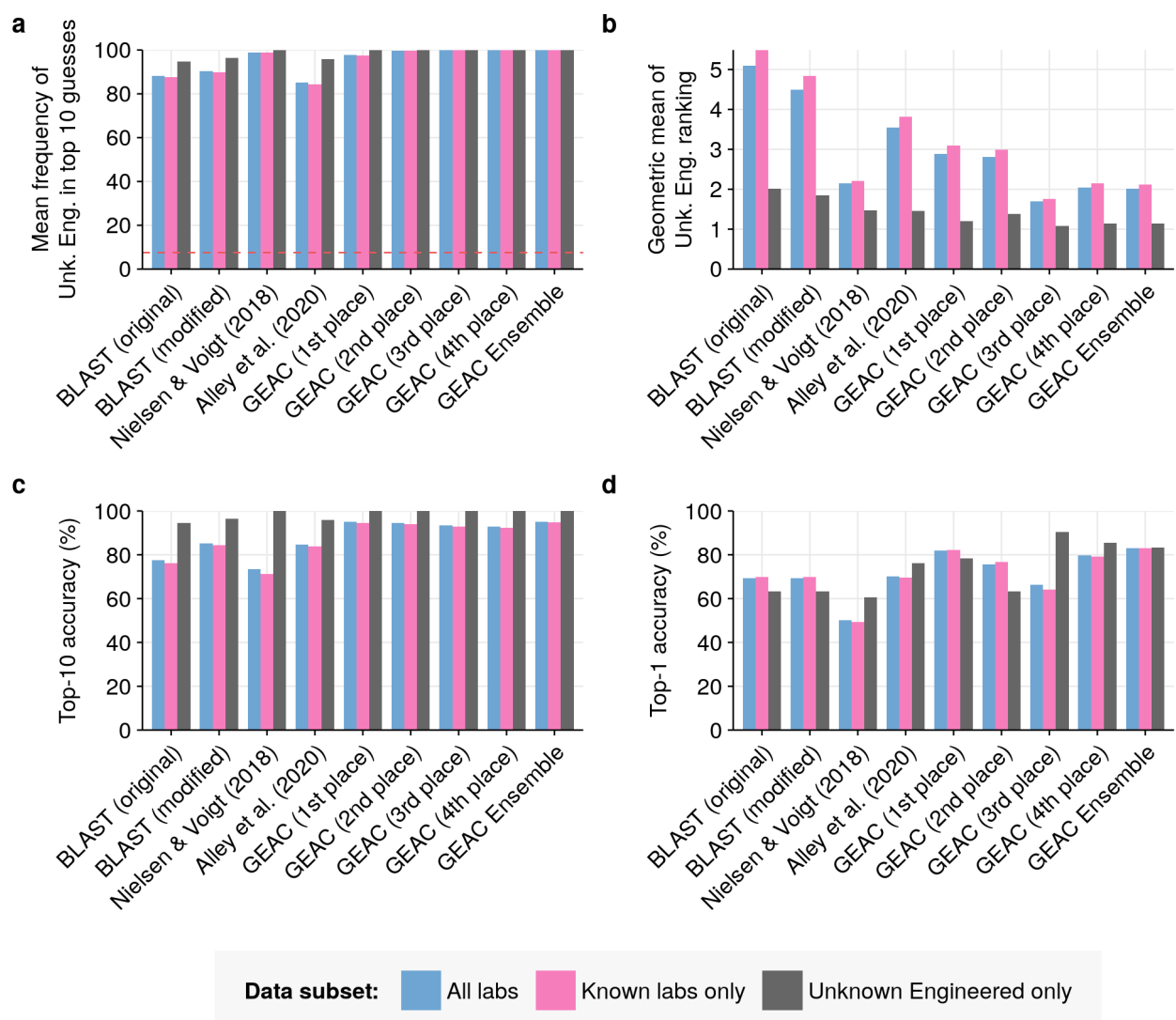

**Supplementary Figure 17: Unknown Engineered metrics for key models.** (a) Mean percentage of sequences in the test set for which key models (BLAST, past ML GEA models, GEA winners and ensemble) included Unknown Engineered in their top-10 lab-of-origin guesses, compared to the true frequency of the category (7.5%, dashed red line). (b) Geometric mean rank of Unknown Engineered, across all sequences in the test set, for each key model. (c) Top-10 accuracy achieved by each key model on each of three subsets of the test set: sequences from all lab categories, sequences from known labs only (excluding Unknown Engineered), and Unknown Engineered sequences only. (d) Top-1 accuracies achieved on the same data subsets.

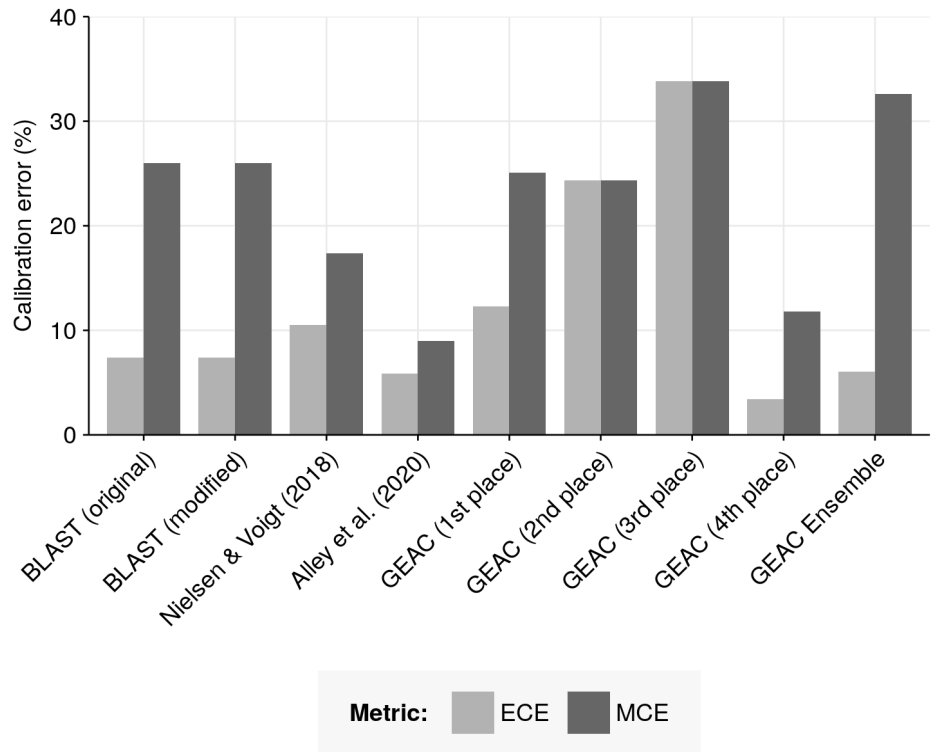

**Supplementary Figure 18: Calibration of key models.** Expected calibration error (ECE) and maximum calibration error (MCE) of Prediction Track winners and ensemble, as compared to BLAST and previously-published ML-based GEA models.

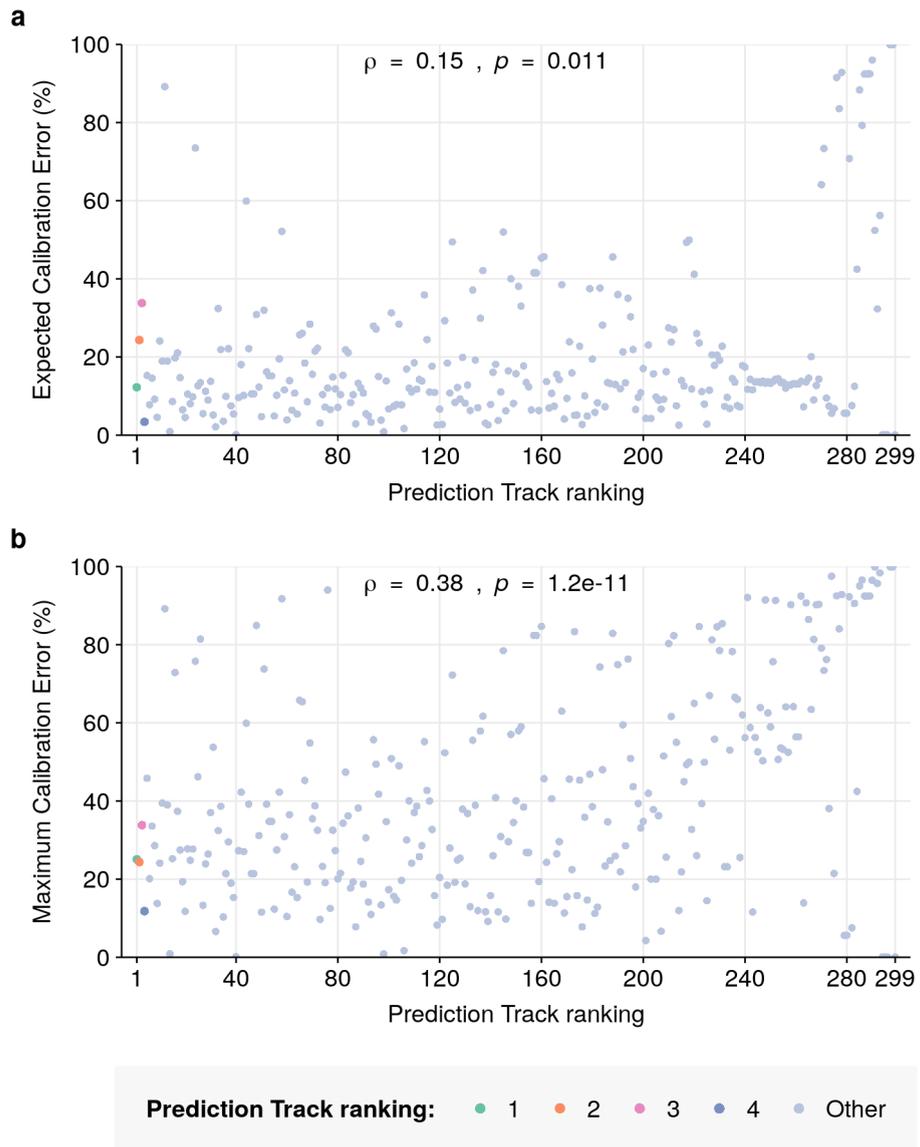

**Supplementary Figure 19: Calibration of Prediction Track teams.** (a) Expected and (b) maximum calibration error of each team in the Prediction Track.

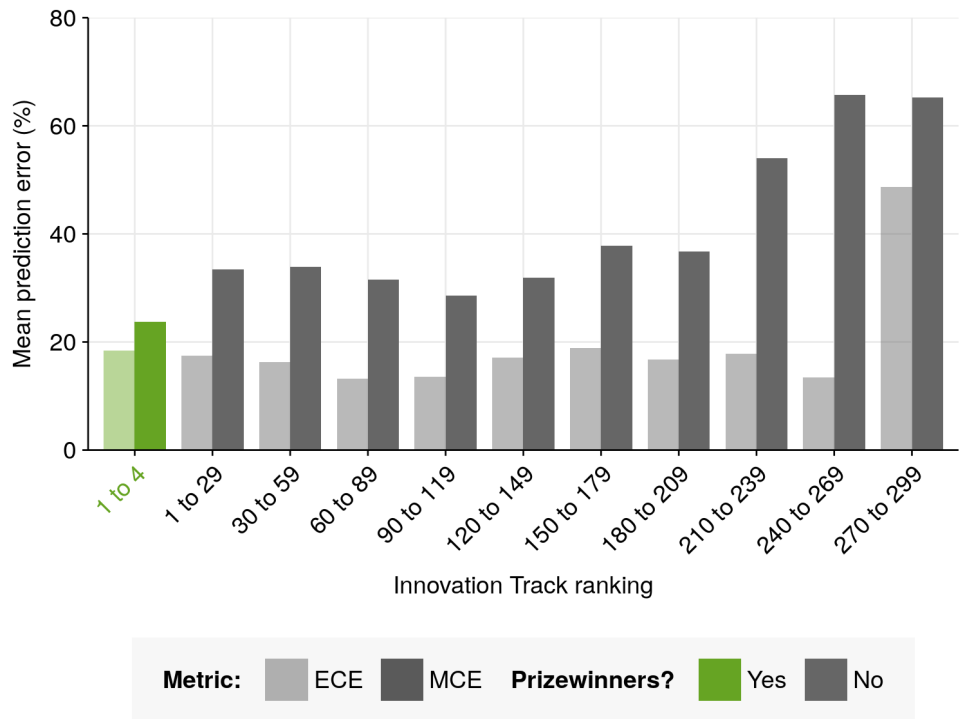

**Supplementary Figure 20: Calibration of Prediction Track teams by decile.** Expected and maximum calibration errors for teams in each rank decile of the Prediction Track (grey bars) as well as for the four winning teams (green bar).

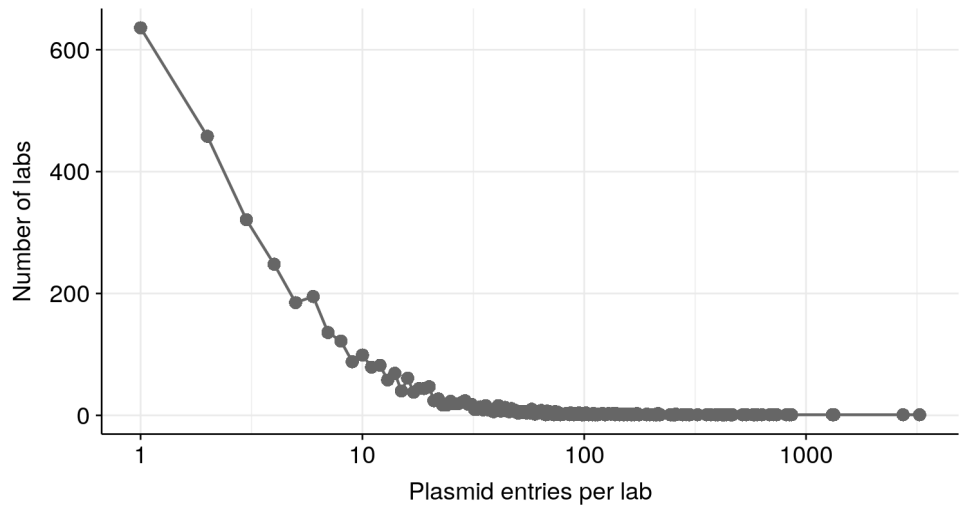

**Supplementary Figure 21: Distribution of plasmid entries per lab in the raw Addgene dataset.** Number of labs with a given number of sequence entries in the Addgene plasmid database, prior to processing and splitting by Alley et al.[2].

**Supplementary Note: Innovation Track problem description**

*What follows is the text of the Innovation Track problem description given to participants at the start of the Genetic Engineering Attribution Challenge. The original document also included a list of judges, which has been omitted here.*

---

# Innovation track

## How would your approach perform in the real world?

### Introduction

In the Innovation Track, we challenge you to bridge the gap between your work in the Prediction Track and real-world attribution problems in academia, industry, and forensics. To succeed, submissions should demonstrate how their lab-of-origin prediction models excel in domains beyond raw accuracy. Submissions will be evaluated by a panel of judges, including world leaders in synthetic biology, data science and biosecurity policy. Winners will receive prizes equal to those provided for the Prediction Track.

### How to compete – beat the benchmark

Any model that achieves a higher top-10 accuracy score than our BLAST benchmark will qualify for entry to the Innovation Track. Teams that exceed this threshold (i.e. **exceeding 75.6% on the private leaderboard**) will automatically be invited to submit a report for assessment by our panel of judges (see "Assessment", below).

If a large number of teams successfully beat the benchmark and make submissions, altLabs and our partners will perform pre-screening, based on the criteria below, to select which submissions to pass to our judges.

### How to win

To succeed in the Innovation Track, you will need to convince experts from a variety of fields that your submission represents valuable progress in solving real-world attribution problems. To do this, you will need to demonstrate – in **plain language** – that your approach is impressive beyond raw lab-of-origin accuracy. This task is fully open-ended: you can write about whatever aspects of your model you think will most impress the judges. Some examples of important criteria you might want to address include:

- **Calibration**: Does your model accurately report its uncertainty about a sample's lab-of-origin?
- **Robustness**: Is your model robust to outliers, changes in the dataset, and unseen labs?
- **Interpretability & biological groundedness**: Explain how your model makes its predictions in a way that would be useful to a biologist or forensic scientist. Do the features it considers important make biological sense? How might you go about integrating the outputs of your model with other biological data sources?
- **Responsible development**: Does your approach thoughtfully address the practical and societal challenges which might arise in real-world applications? Have your design decisions been made with diverse stakeholders in mind? How might you continue the responsible development of your approach in the future?

This list is not exhaustive; we expect a wide variety of approaches to do well in the Innovation Track. The key is to demonstrate that your model brings us closer to concrete application.

**Note on data use**: External data may be used in the Innovation Track for the purposes of showcasing the capabilities of the model post-training. External data may not be used for model training or for any use in the Prediction Track.

If you need more inspiration, we've provided three example scenarios below. We emphasize that these are just examples; you do not need to address these specific scenarios to succeed.

**Examples**

**Example 1: Detecting genetic plagiarism in a multi-source plasmid**

The [International Genetically Engineered Machine (iGEM) competition](#) is the world's premiere student competition in synthetic biology, in which teams of students work together to design, build, test, and measure systems of their own design using interchangeable biological parts. As part of their competition submissions, iGEM teams must provide detailed information regarding the origin of ideas and components they use. With thousands of submissions to review, detecting and investigating misattributed ideas is becoming a challenge for iGEM.

As part of their submission, one iGEM team presents [composite parts](#) of their own design, containing sequences derived from multiple different sources. Some of these parts are from the iGEM parts registry, while others are claimed to be of their own design. Could your model identify which components of each plasmid come from which sources, and thus help assess whether all of the team's work have been correctly attributed?

**Example 2: Accidental release**

A sewage treatment plant notices a strange, unsettling green glow in the...sludge. They send a sample to a local microbiology lab, where it's classified as a [*E. coli*](#), probably the most common laboratory microbe. While harmless, the bacteria have been genetically engineered to express [green fluorescent protein](#): someone is being sloppy with biowaste disposal.

Authorities in the city turn to you to find out which lab is responsible; however, many of the labs in the city are not in your database. In the event that a precise identification is not possible, could your model provide any partial information that would help human investigators narrow the search?

**Example 3: Synthesis Screening**

A (fictional) DNA synthesis company has a serious problem with fraud: they offer large discounts to academic institutions, but suspect that many private startups are using academic contacts to exploit this, potentially losing millions of dollars in revenue. Given the immense volume of orders, it is impossible to manually inspect each sequence to verify its purported lab-of-origin. So, the company has turned to attribution models to automate verification.

To be useful and economical in this context, an attribution model must overcome certain challenges: among others, it must be very computationally cheap to run, and perform well given only DNA sequence fragments (no phenotype information). Most importantly, to minimize disruption to legitimate business, it should have a very low false-positive rate, while maintaining an acceptably low rate of false negatives. Could your model fulfill these criteria?

**Assessment**

At the close of the Prediction Track on October 19th, any team whose accuracy exceeds the BLAST benchmark on the **private leaderboard** will be invited to submit a report on their approach. This report should be at most four pages long, with at most two figures, and should:

- Demonstrate a **novel and creative** approach to genetic engineering attribution;
- Demonstrate what capabilities of their model, other than raw accuracy, would make it useful for solving **realistic attribution problems**;
- Show thoughtful consideration of the **societal impact and use** of attribution models;
- Discuss the **limitations** of the model, and how further work might improve it;
- Be **comprehensible** to both technical and non-technical readers.

Assessment of submissions to the Innovation Track will be conducted by a diverse panel of distinguished judges, including top experts in synthetic biology, data science, and biosecurity policy. Each submission will be assessed by multiple judges. As such, to succeed in the Innovation Track, your report should be **comprehensible to both technical and non-technical readers.** Innovation Track prize-winners will be announced alongside those for the Prediction Track.

---

**FAQ**

**Who can submit to the Innovation Track?** Any team that beats our BLAST benchmark on the Prediction Track can submit a report to the Innovation Track.

**What is the exact benchmark value we need to beat?** The BLAST benchmark for top-10 accuracy is 78.8% on the public leaderboard and 75.6% on the private leaderboard. To qualify for the Innovation Track, your model should score better than 75.6% on the private leaderboard.

**What if we have multiple submissions to the Prediction Track?** Each team may make at most one submission to the Innovation Track. If your team makes multiple submissions, you must select one to discuss in your Innovation Track entry.

**Nobody on our team is a biologist! Can we compete?** Any team that beats the BLAST benchmark in the Prediction Track is welcome to compete in the Innovation Track. Collaborating with biology or policy specialists may be helpful for this track, but is not required.

**How long will we have to prepare our report?** Following closure of the Prediction Track on October 19th 2020, notified participants will have up to two weeks to prepare their submissions to the Innovation Track. However, teams that are confident they've beaten the benchmark are strongly encouraged to prepare their submissions well in advance. The submission window for the Innovation Track will close at 11:59pm (UTC) on November 1st, 2020.

**What should I submit?** Participants in the Innovation Track should submit the following:

- The code for your lab-of-origin prediction model. This should be the same model you used in your submission to the Prediction Track.
- Your report, which should explain in plain language how your model fulfills the criteria outlined above.
- Code and data (see below) for generating any figures you include in your report.

The code and data files do not count towards the four-page limit for the report.

**How should the report be formatted?** Reports submitted to the Innovation Track should be written in English, in PDF format, at most four pages long (Letter or A4), and with at most two figures. There is no minimum length requirement. **To enable blind assessment of submissions, your report should not include the names of your team or team members.** There are no other detailed formatting requirements, but the report should conform to the norms of a good scientific paper: intellectual contributions from outside your team should be acknowledged and cited, and methods should be reported transparently.

**What data can I use for the model I submit to the Innovation Track?** The model you submit to the Innovation Track should be the same as the corresponding submission to the Prediction Track. As such, **the model should be trained only on the dataset provided for this competition**. External data may also be used in the Innovation Track for the purposes of showcasing the capabilities of the model post-training. As with the Prediction Track, finalists for the Innovation Track will have their model performance validated against an out-of-sample verification set; teams judged to have violated rules regarding data usage will be disqualified.

**What data can I use for the figures in my Innovation Track report?** While the model submitted to the Innovation Track should be trained only on the data provided for the Prediction Track, you are welcome to use other data to illustrate the capabilities of that model in your Innovation Track report. For example, if your team submits a report based on Example 1 above, you would be welcome to acquire or design some multi-source plasmids to demonstrate your model's ability to distinguish subsequences from different sources. Any additional data you use in this way must be included in your submission to the Innovation Track, alongside the code for your model and figures.

**Who owns the intellectual property for my Innovation Track submission? What happens to my model after the competition?** As with the Prediction Track, the IP for all prize-winning submissions will be assigned to altLabs upon receipt of prize money. After the competition, altLabs will seek input from various stakeholders – including prizewinning teams – on how best to use these results to promote responsible innovation. altLabs is a non-profit organisation, and will never sell or otherwise monetize prize-winning submissions. The IP for submissions that do not win prizes remains with their respective teams.